\documentclass[lettersize,journal]{IEEEtran}
\usepackage{amsmath,amsfonts}
\usepackage{algorithmic}
\usepackage{algorithm}
\usepackage{array}
\usepackage[caption=false,font=normalsize,labelfont=sf,textfont=sf]{subfig}
\usepackage{textcomp}
\usepackage{orcidlink}
\usepackage{stfloats}
\usepackage{url}
\usepackage{verbatim}
\usepackage{graphicx}
\usepackage{cite}
\newcommand{\etal}{\textit{et al. }}

\newcommand{\myfigref}[1]{Fig.~\ref{#1}}
\newif\iftrackversion
  \newcommand{\revisetypo}[1]{#1}
  \newcommand{\reviseone}[1]{#1}
  \newcommand{\revisetwo}[1]{#1}
  \newcommand{\revisethree}[1]{#1}
\iftrackversion

\else
  
\fi
\usepackage{multirow}
\usepackage{booktabs}
\begin{document}
\hypersetup{hidelinks,
	colorlinks=true,
	allcolors=black,
	pdfstartview=Fit,
	breaklinks=true}
\title{MLG-Stereo: ViT Based Stereo Matching with Multi-Stage Local-Global Enhancement}

\author{Haoyu Zhang~\orcidlink{0009-0007-9819-311X}, Jingyi Zhou~\orcidlink{0009-0005-6967-0090}, Peng Ye, Jiakang Yuan, Lin Zhang\\ Feng Xu,~\IEEEmembership{Senior Member,~IEEE,}  Tao Chen~\orcidlink{0000-0002-0779-9818}, ~\IEEEmembership{Senior Member,~IEEE,} 
\thanks{Resubmitted on March 26, 2026. Revised on February 15, 2026. Initially submitted on April 13, 2025. This work is supported by National Key Research and Development Program of China (No. 2022ZD0160101), Shanghai Natural Science Foundation (No. 23ZR1402900), Shanghai Science and Technology Commission Explorer Program Project (24TS1401300), Shanghai Municipal Science and Technology Major Project (No.2021SHZDZX0103). The computations in this research were performed using the CFFF platform of Fudan University. (Haoyu Zhang and Jingyi Zhou contributed equally to this work) (Corresponding author: Peng Ye and Tao Chen.)}
\thanks{Haoyu Zhang, Jingyi Zhou, Jiakang Yuan, Lin Zhang and Tao Chen are with the Embedded Deep Learning and Visual Analysis Laboratory, College of Future Information Technology, Fudan University, Shanghai 200433, China (e-mail: 25113090081@m.fudan.edu.cn; zhoujingyi19@fudan.edu.cn; jkyuan22@m.fudan.edu.cn; 22110720068@m.fudan.edu.cn; eetchen@fudan.edu.cn)}
\thanks{Peng Ye is with the Embedded Deep Learning and Visual Analysis Laboratory, College of Future Information Technology, Fudan University, Shanghai 200433, China, and also with Shanghai Artificial Intelligence Laboratory, Shanghai 200232, China, and also with the Chinese University of Hong Kong, Hong Kong (e-mail: 20110720039@fudan.edu.cn)}
\thanks{Feng Xu is with the Key Laboratory for Information Science of Electromagnetic Waves, Ministry of Education, Fudan University, Shanghai 200433, China, and also with the College of Future Information Technology, Fudan University, Shanghai 200433, China (e-mail: fengxu@fudan.edu.cn) }
}

\markboth{Journal of \LaTeX\ Class Files,~Vol.~14, No.~8, August~2021}%
{Shell \MakeLowercase{\textit{et al.}}: A Sample Article Using IEEEtran.cls for IEEE Journals}

\IEEEpubid{0000--0000/00\$00.00~\copyright~2021 IEEE}

\maketitle

\begin{abstract}
With the development of deep learning, \revisetypo{ViT-based stereo matching methods have made significant progress due to their remarkable robustness and zero-shot ability}. However, due to the limitations of \revisetypo{ViTs in handling resolution sensitivity and their relative neglect of local information}, the ability of ViT-based methods to predict details and \revisetypo{handle arbitrary-resolution images} is still weaker than that of CNN-based methods. To address these shortcomings, we propose MLG-Stereo, a systematic pipeline-level design that extends global modeling beyond the encoder stage. First, we propose a Multi-Granularity Feature Network to effectively balance global context and local geometric information, enabling comprehensive feature extraction from images of arbitrary resolution and bridging the gap between training and inference scales. Then, a Local-Global Cost Volume is constructed to capture both \revisetypo{locally-correlated} and global-aware matching information. Finally, a \revisetypo{Local-Global Guided Recurrent Unit} is introduced to iteratively optimize the disparity \revisetypo{locally} under the guidance of global information. Extensive experiments are conducted on multiple benchmark datasets, demonstrating that our MLG-Stereo exhibits highly competitive performance on the Middlebury and KITTI-2015 benchmarks compared to contemporaneous leading methods, and achieves outstanding results in the 
KITTI-2012 \revisetypo{dataset}.
\end{abstract}


\section{Introduction}

Stereo matching seeks to estimate dense pixel-wise correspondences—specifically, disparity—between rectified image pairs captured by horizontally displaced cameras. Due to its \revisetypo{high accuracy in depth estimation enabled by} camera calibration, stereo matching has established itself as a cornerstone task in computer vision and 3D reconstruction. Its practical relevance spans diverse real-world domains such as robotics, autonomous driving, and augmented reality~\cite{fan2020computer,yao2023re,bi2023application,samadi2013new,nie2019exploring, cheng2024stereo}, where accurate scene depth perception is indispensable for \revisetypo{environment modeling, intelligent decision-making, and immersive human-computer interactions}.

\begin{figure}[t]
  \centering
   \includegraphics[width=0.95\linewidth]{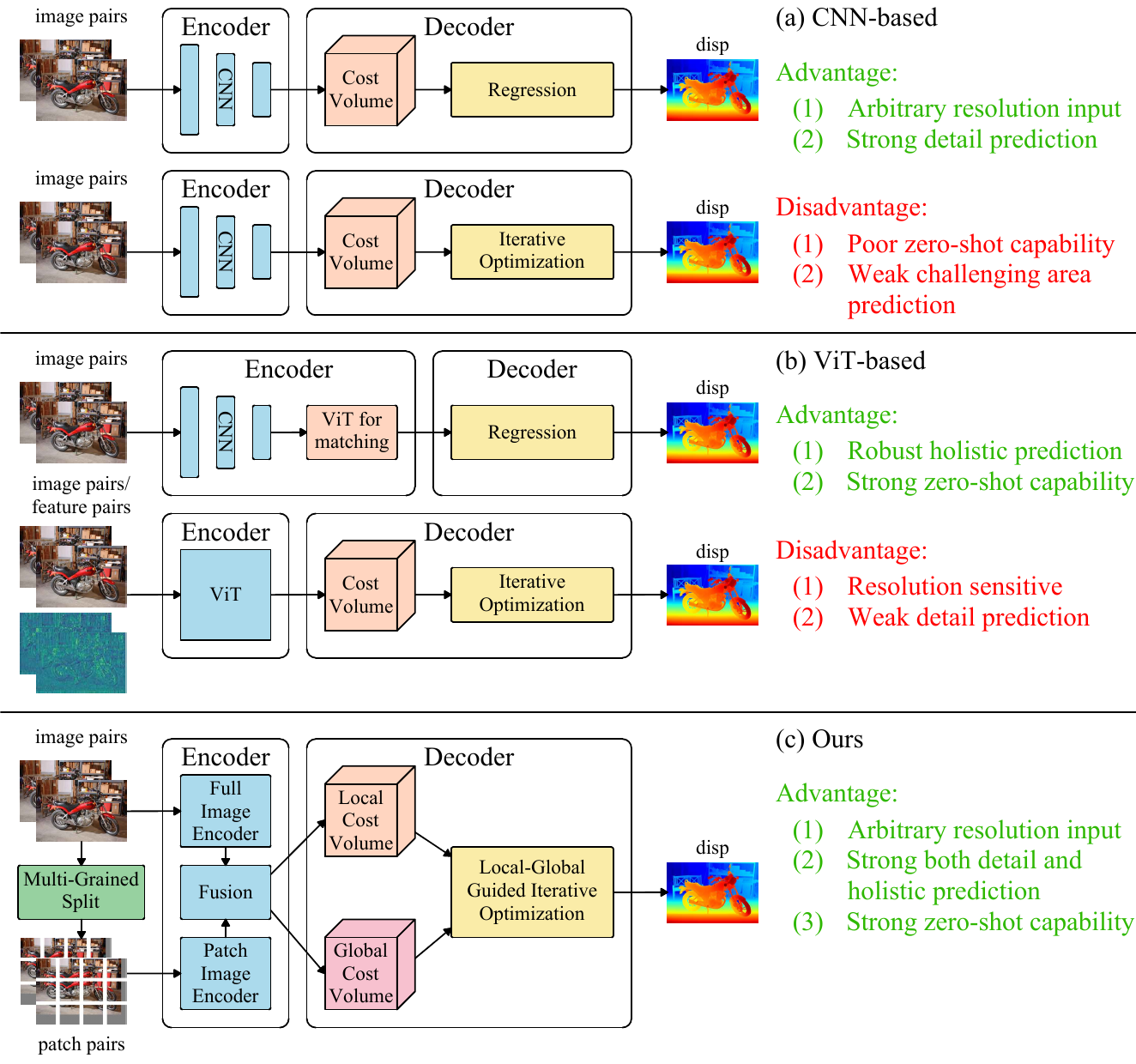}
   \caption{Overview of existing typical stereo matching frameworks. (a) CNN-based frameworks. (b) ViT-based frameworks. (c) Our MLG-Stereo framework which builds on the ViT-based method and improves it with \revisetypo{multi-stage} local-global enhancement. This design allows our method to have the advantages of both CNN-based and ViT-based methods.
   }
   \label{fig:fig_1}
\end{figure}

With the development of deep learning, Convolutional Neural Networks (CNNs) have been applied to stereo matching, significantly improving accuracy. Early works mainly consist of aggregation-based methods~\cite{gcnet,psmnet,gwcnet}, as shown in the upper part of~\myfigref{fig:fig_1}(a), where CNNs acted as encoders to extract features and 3D convolutions were used to construct a 3D cost volume, followed by regression to predict disparity. Since RAFT-Stereo~\cite{raft-stereo} introduced the idea of iterative optimization into the stereo matching task, many researchers have devised various methods~\cite{zhao2023high,zhao2022eai,igevstereo,wang2024selective,zhou2024all}, as shown in the lower part of~\myfigref{fig:fig_1}(a). These methods typically use a multi-scale CNN encoder that includes a feature network and a context network, followed by \revisetypo{initializing} disparity using regression \revisetypo{or zero disparity initialization}, and decoding residual disparity using Gated Recurrent Unit (GRU) or Long Short-Term Memory (LSTM) based \revisetypo{blocks}. In general, these CNN-based methods not only support arbitrary resolution input but also bring stronger local \revisetypo{detail prediction} ability through local perception. However, these methods have poor zero-shot capabilities due to the limited general feature extraction ability of CNN networks, and their prediction errors are prone to occur in some difficult areas (such as repeated textures and weak textures) due to the lack of a global receptive field.\par
\IEEEpubidadjcol

Inspired by the powerful global context extraction capability of Vision Transformers (ViTs), ViT-based methods have been explored in recent years. Some researchers adopt a hybrid approach by combining CNNs and ViTs in a sequential manner, where CNNs serve as feature encoders and ViTs are used for matching, such as STTR~\cite{li2021revisiting} and CSTR~\cite{guo2022context}, as shown in the upper part of~\myfigref{fig:fig_1}(b). Some researchers, on the other hand, utilize the powerful feature extraction capability of ViT to directly extract image features or enhance the features extracted by CNNs, as shown in the lower part of~\myfigref{fig:fig_1}(b)~\cite{xu2023unifying,li2024roadformer}. These ViT-based methods rely on the attention mechanism to make more robust global receptive field image predictions and can leverage Vision Foundation Models (VFMs) as the backbone to achieve stronger zero-shot capabilities. However, ViTs are highly sensitive to resolution, resulting in degraded performance on ultra-high-resolution images. Moreover, their limited ability to capture fine local details often results in detail loss, as observed in the results of VITAStereo~\cite{li2024roadformer} and our ablation experiments.
In addition to the encoders, current decoder architectures also face challenges in fully leveraging the global matching information extracted by the encoder. Due to the computational burden of the cost volume, existing methods typically feed only local window cost volume information into the iterative optimization process. The underutilization of global matching information in existing methods makes the optimization process prone to local optima and often requires more iterations to achieve satisfactory stereo matching results.

\reviseone{To address these challenges, we propose MLG-Stereo, a unified pipeline for ViT-based stereo matching. We observe that while CNN-based stereo matching inherently excels at preserving local details, its global modeling capacity is constrained by local perception fields. Conversely, ViT-based methods demonstrate robust global context understanding but face bottlenecks in arbitrary-resolution adaptation and fine detail recovery. Consequently, the core design philosophy of our work is to amplify the potential of ViT-based stereo matching by overcoming their inherent limitations, which systematically couples global geometric guidance with local iterative refinement. Specifically, we first propose a Multi-Granularity Feature Network (MGFN) that employs a shared-weight hierarchical architecure to overcome single-scale constraints, bridging pretrained global semantics with local fine-grained geometric details. Building on this, we introduce the Local-Global Cost Volume (LGCV), leveraging latent token compression to achieve bidirectional global cost aggregation within hardware-affordable limits. Finally, we design a Local-Global Guided Recurrent Unit (LGRU) that genuinely injects global cost priors into iterative updates, rectifying the optimization direction in ill-posed regions.} As a result, our method effectively balances global context and local fine-grained information. Extensive evaluations on multiple public benchmarks demonstrate the superiority of our framework. Notably, our method reached leading performance on the Middlebury and KITTI-2015 benchmarks, while securing highly competitive results on 
KITTI-2012.

Our main contributions can be summarized as follows:
\begin{itemize}
  \item \reviseone{We introduce MLG-Stereo, a framework that unlocks the full potential of ViT-based stereo matching. It extends local-global enhancement to all stages of ViT-based pipelines, and natively supports arbitrary input resolutions, delivers both fine-grained detail and holistic scene prediction, and exhibits strong zero-shot generalization.}
  \item \reviseone{We design a Multi-Granularity Feature Network (MGFN) that overcomes the fixed-resolution constraint of VFMs. By fusing features from multi-scale patches with global contexts, it enables precise feature extraction for inputs of arbitrary scales.}
  \item \reviseone{We introduce a Local-Global Cost Volume (LGCV) and a Local-Global Guided Recurrent Unit (LGRU). To address memory bottlenecks in global matching, LGCV utilizes latent tokens to condense broad correlation information. LGRU then leverages this global guidance to rectify local optimization, enhancing robustness in ill-posed regions.}
  \item Extensive evaluations on public benchmarks demonstrate the superiority of our structural design. Our robust enhancement results in leading performance on Middlebury and KITTI-2015 datasets, and demonstrates substantial performance gains on 
  KITTI-2012 dataset over recent competitive baseline models.
\end{itemize}

\section{Related Work}
\subsection{Stereo Matching}
As a widely used algorithm, stereo matching has been studied for a long time. Early algorithms~\cite{boykov2001fast,klaus2006segment,sun2003stereo,yang2008stereo,hirschmuller2002real,van2002hierarchical,hirschmuller2005accurate} use hand-crafted features and priori rules to generate disparity. However, there is still a lot of room for improvement in the accuracy of these traditional methods.

\subsubsection{CNN-based Methods}
Since Zbontar and LeCun~\cite{zbontar2015computing} introduced CNNs to compute the matching cost, learning-based algorithms have progressively replaced traditional stereo matching methods. GCNet~\cite{gcnet}  leverages 3D CNN to jointly capture geometric constraints and contextual information, effectively addressing challenges in occluded and textureless regions. PSMNet~\cite{chang2018pyramid} integrates global context information through a Spatial Pyramid Pooling (SPP) module and stacked hourglass 3D CNN modules, enhancing the accuracy of matching. GWCNet~\cite{gwcnet} proposes group-wise correlation cost volume to reduce the cost of building cost volume and improve the accuracy of prediction. In recent years, iterative optimization-based algorithms have gradually become mainstream. Inspired by RAFT~\cite{raft}, RAFT-Stereo~\cite{raft-stereo} first explores the iteration of multi-scale GRU-based update blocks to generate the final disparity map by iterating. IGEV-Stereo~\cite{xu2023iterative} extracts the missing geometric information in the All Pair Correlation(APC) to an additional Geometry Encoding Volume(GEV) and applies it to iterative optimization. Selective-Stereo~\cite{wang2024selective} proposes a Contextual Spatial Attention module(CSA) and a Selective Recurrent Unit(SRU) to better preserve high-frequency detail information during iteration. As a leading method, AIO-Stereo~\cite{zhou2024all} enhances the feature extraction capability of the encoder using a strategy of distilling from multiple VFMs and selectively forward fusion. \revisethree{Although these methods excel in local detail prediction, their localized perception fields primarily focus on local contexts, which leaves open the challenge of effectively handling ill-posed regions with complex global structures.}
\par
\subsubsection{ViT-based Methods}
More recently, Vision Transformers(ViT) have been widely used in various computer vision tasks due to their global feature extraction capability. And stereo matching is no exception, researchers apply ViT in stereo matching model from different perspectives. On the one hand, STTR~\cite{li2021revisiting} and CSTR~\cite{guo2022context} treat the stereo matching task as a sequence-to-sequence correspondence task, and use ViT instead of the construction of cost volume to directly generate the predicted disparity. \revisetypo{On the other hand}, GMStereo~\cite{xu2023unifying} uses ViT to enhance image features globally within and between images, giving the model the ability to infer large disparity scenarios with fewer iterationss. In addition, VITAStereo~\cite{li2024roadformer} directly uses ViT as the backbone, achieving high-precision disparity prediction. 
However, current ViT-based stereo matching models primarily focus on global modeling at the encoder stage. They often encounter bottlenecks when adapting to arbitrary high-resolution images, and their reliance on global tokens occasionally compromises local fine-grained information compared to CNN-based method's detail preservation.

\revisetwo{Recently, works such as FoundationStereo~\cite{wen2025foundationstereo}, DEFOM-Stereo~\cite{jiang2025defom}, and Monster~\cite{cheng2025monster} have begun to explore hybrid architectures utilizing ViT-based VFMs combined with side-tuning CNNs. These methods attempt to mitigate the shortcomings of ViTs in processing high-resolution images and making fine-grained predictions by employing parallel CNN branches. While these methods bridge the gap effectively, their global modeling mostly stays in the encoder. This leaves open the question of how to extend systematic local-global integration consistently across the entire stereo matching pipeline, which motivates our proposed structural design.}

\par
\subsection{Vision Foundation Models and their downstream applications}
\subsubsection{Vision Foundation Model}
In recent years, advancements in hardware performance and the availability of large-scale datasets have spurred the emergence of high-performance vision foundation models (VFMs). These models excel at processing and understanding image or video data, significantly driving progress in various vision tasks. Swin family~\cite{liu2021swin,liu2021swinv2}, which adopts a hierarchical structure similar to that of the CNN network and a shifted window-based self-attention. DINO family~\cite{caron2021emerging,oquab2023dinov2} leverages self-supervised learning on vision transformers to enhance performance. 
\subsubsection{Applications in Downstream Task}
Due to the powerful performance of VFM, researchers have widely applied them to various downstream tasks. Swin-unet~\cite{cao2022swin} adopts the hierarchical Swin Transformer as its backbone within a U-shaped architecture, improving the performance of medical image segmentation, particularly effective in polyp segmentation tasks. DepthAnything family~\cite{depth_anything_v1,depth_anything_v2} greatly improves the performance of monocular depth estimation by utilizing DINOv2's powerful feature extraction capability and a large amount of unlabeled data. Depth Pro~\cite{yang2024depth} has designed a multi-scale ViT for fixed size images, greatly improving the performance of monocular depth estimation models in detail areas. However, due to the fact that stereo matching estimation models not only need to focus on global and local information, but also need to support inputs of various resolutions, there is still a lot of room for improvement in the performance of current stereo matching models using VFM.

\section{Method}
\begin{figure*}[t]
  \centering
   \includegraphics[width=0.95\linewidth]{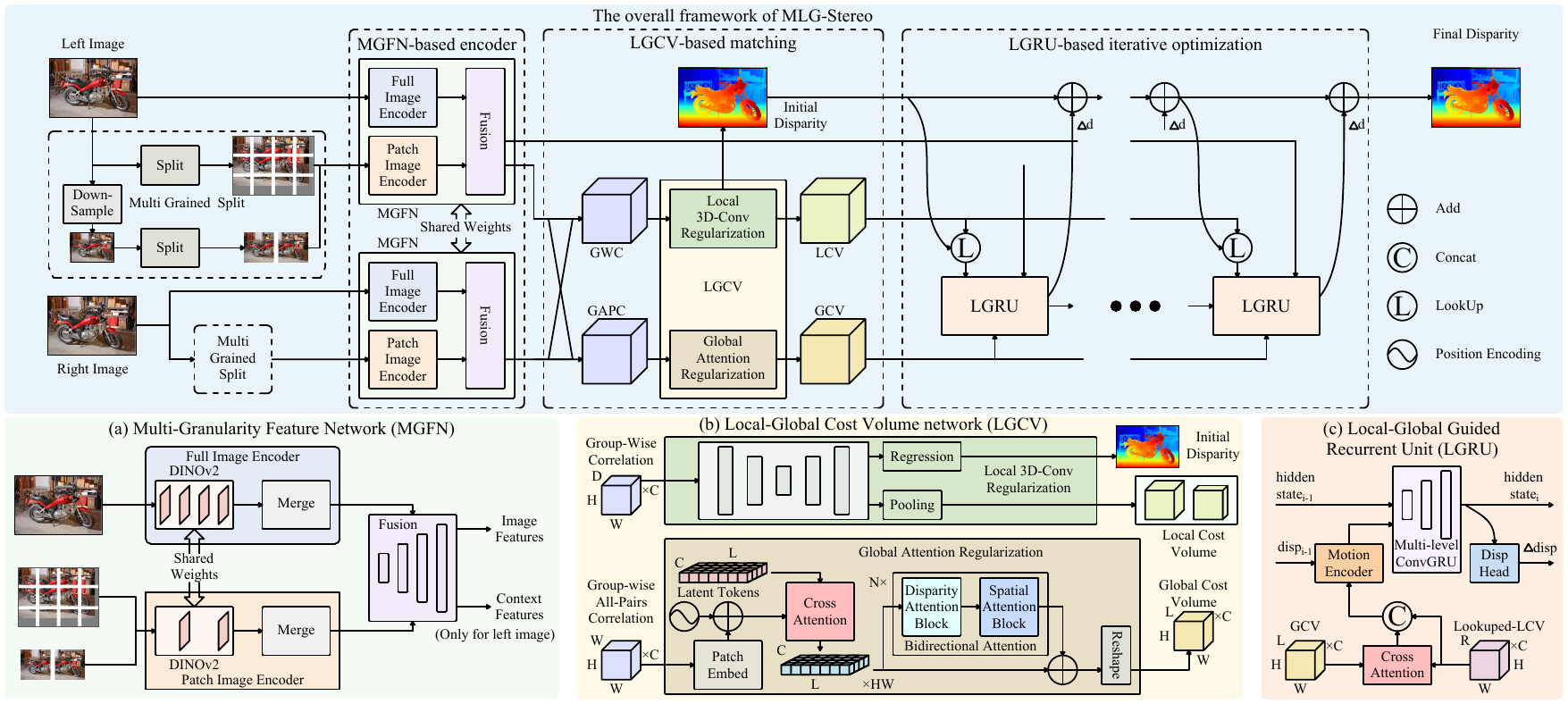}
   \caption{The overall framework of our MLG-Stereo. The workflow of MLG-Stereo is to provide a pair of images, first extract the features of the images using an encoder containing MGFN, then generate local and global matching information by LGCV matching block, and finally use LGRU to refine local disparity guided by global matching information. (a) Multi-Granularity Feature Network (MGFN) (b) Local Global Cost Volume (LGCV) matching block. (c) Local-Global Guided Recurrent Unit (LGRU).}
   \label{fig:fig_2}
\end{figure*}

\revisethree{To address the critical limitations of current ViT-based stereo matching methods, including sensitivity to input resolution, loss of fine local details, and the restriction of global enhancement to the feature encoder, we propose MLG-Stereo. This novel ViT-based stereo matching framework introduces multi-stage local-global enhancement, extending the powerful global context modeling of ViTs across the entire stereo matching pipeline. As illustrated in~\myfigref{fig:fig_2}, MLG-Stereo incorporates a Multi-Granularity Feature Network (MGFN) to  capture both fine local details and high-level global semantics using VFMs. Additionally, it features a Local-Global Cost Volume (LGCV) to construct both local and global cost volumes, and a Local-Global Guided Recurrent Unit (LGRU) to enable globally-guided iterative optimization.}

\subsection{Multi-Granularity Feature Network-based Feature Encoder}

\revisethree{Recently, researchers have explored directly substituting traditional backbones with ViT-based VFMs in stereo matching tasks (e.g., VITAStereo~\cite{li2024roadformer}). This paradigm brings notable benefits, significantly enhancing the global feature extraction capability to capture broad contextual priors. However, directly applying VFMs encounters critical challenges in high-resolution fine-grained prediction. Specifically, to handle the varying resolutions of input images (e.g., MiddEval v3), resizing them to a fixed resolution is commonly adopted to bridge the discrepancy between training and inference. Yet, this inevitably results in a severe loss of local geometric details and significant performance degradation when testing on high-resolution images. To overcome this limitation, our design philosophy is to fully exploit the powerful feature extraction capabilities of VFMs while preserving intricate details through local enhancement. Inspired by DepthPro~\cite{bochkovskii2024depth}, we propose a Multi-Granularity Feature Network (MGFN) based on the DINOv2 family~\cite{oquab2023dinov2}. By processing both the global full images and the local patched images, MGFN extracts multi-granularity features that effectively unify local fine-grained details with global contextual information.}

Given a pair of left and right images \( I_{l}, I_{r} \in \mathbb{R}^{3 \times H \times W} \), we first divide them into patches at both full and half resolutions, with a patch size of \( 224 \times 224 \) corresponding to the original training input resolution of the chosen backbone. Thus, for each image, two patch sequences are yielded: \( P_{0} \in \mathbb{R}^{N_{0} \times 3 \times 224 \times 224} \), derived from the full-resolution image, and \( P_{1} \in \mathbb{R}^{N_{1} \times 3 \times 224 \times 224} \), obtained from the half-resolution image, where \( N_{0} \) and \( N_{1} \) denote the respective numbers of patches. The overlap rates of the divided patches in the two sequences are 25\% and 50\% respectively.

Then, the two patch sequences, \( P_{0} \) and \( P_{1} \), are fed into the Patch Image Encoder, which employs DINOv2~\cite{oquab2023dinov2} as the backbone network. For each sequence, feature tokens are extracted at two levels, specifically after the middle and last blocks. To refine the features, we remove padding and overlapping tokens and merge them into two-dimensional feature lists: \( \mathcal{T}_{0,l} \) (\( \mathcal{T}_{0,r} \)) at a resolution of \( 1/16 \), and \( \mathcal{T}_{1,l} \) (\( \mathcal{T}_{1,r} \)) at a resolution of \( 1/32 \).

On the other hand, to capture global information from the entire image, the full image pairs \( I_{l}, I_{r} \) are directly fed into the Full Image Encoder, which shares weights with the Patch Image Encoder. Thus, four levels of feature tokens are extracted and subsequently merged into a two-dimensional feature list, \( \mathcal{T}_{2,l} \) (\( \mathcal{T}_{2,r} \)), at a resolution of \( 1/16 \).
 \begin{figure}[t]
  \centering
   \includegraphics[width=1.0\linewidth]{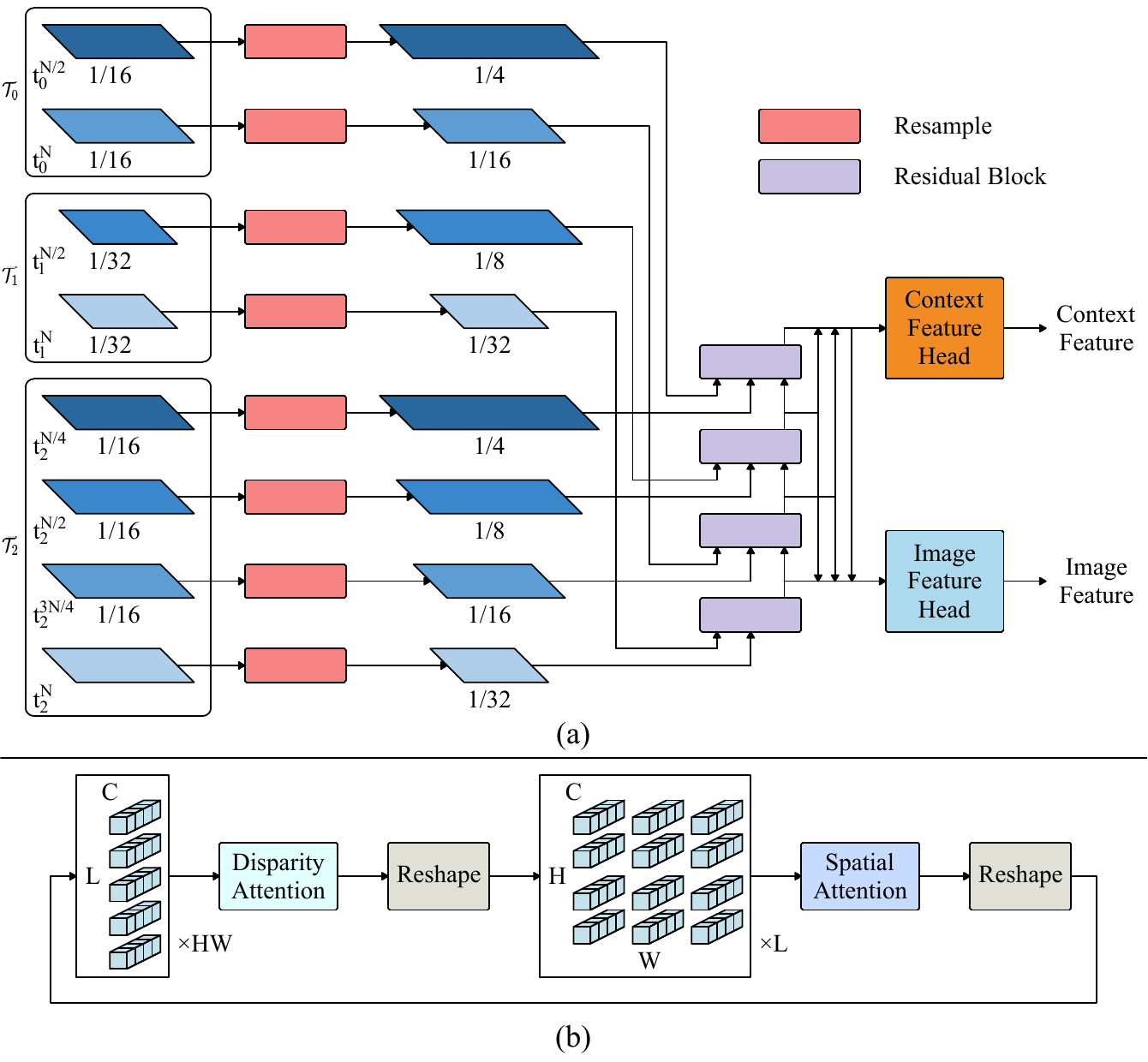}
   \caption{(a) The Feature Fusion Network in MGFN. We use convolution and deconvolution to transform the resolution of features, and then fuse the features using a residual network to generate a feature pyramid, finally two types of feature heads are used to generate image features and contextual features. (b) The pipeline of Bidirectional Attention in LGCV-based matching block. We first use reshape and then conduct attention to GAPC sequences in two directions to extract matching information in both disparity and spatial dimensions.}
   \label{fig:fig_3}
\end{figure}

Utilizing the Feature Fusion Network, we construct the final feature pyramid \( \mathcal{F}_l \ (\mathcal{F}_r) = \{ f_{1},\ f_{2},\ f_{3},\ f_{4} \} \). And finally, an image feature head and a context feature head are employed to generate image features \( f_{l,i} \ (f_{r,i}) \in \mathbb{R}^{dim \times \frac{H}{2^{i+1}} \times \frac{W}{2^{i+1}}} \) for \( i = 1,2,3,4 \), and context features \( f^c_{l,i} \in \mathbb{R}^{dim \times \frac{H}{2^{i+1}} \times \frac{W}{2^{i+1}}} \) for \( i = 1,2,3 \), as illustrated in ~\myfigref{fig:fig_3}(a). Furthermore, to better adapt to the stereo matching task, we fine-tune all normalization layers and the last several blocks in the DINOv2 backbone network.

\subsection{Local Global Cost Volume-based Matching Block}
\revisethree{Similarly, in the cost volume aggregation phase, current ViT-based architectures frequently adopt the 3D CNN strategies introduced by Xu \etal~\cite{igevstereo}. This methodology yields significant benefits by supplementing the network with crucial non-local geometric and contextual information. However, aggregating the cost volume via 3D CNNs inherently restricts the receptive field. This limitation hinders the model's ability to capture long-range dependencies, often leading to matching ambiguities in challenging regions such as textureless areas or occlusions. To resolve this, our design philosophy aims to leverage attention mechanisms for comprehensive global aggregation while strictly constraining the memory overhead. Consequently, we design a Local-Global Cost Volume (LGCV) module. While retaining local aggregation to preserve fine-grained matching details, our LGCV innovatively utilizes latent tokens to compress the cost volume and perform efficient global attention aggregation, thereby achieving robust and accurate disparity estimation without incurring unacceptable computational costs.}

Following IGEV-Stereo, we construct the LCV by locally regularizing the Group-Wise Correlation (GWC) via a 3D hourglass network and pooling layers, which integrates multi-level local matching information. To further capture long-range dependencies, we introduce a Global Attention Regularization Network to generate the GCV, thereby extracting global matching information. \par

Specifically, we first apply group-wise correlation (setting the number of groups to 8 in our method) to the 1/4 resolution feature pairs $f_{l,1}$ and $f_{r,1}$ to obtain Grouped All-Pairs Correlation (GAPC):
\begin{equation}
\begin{aligned}
  &Corr(g,z,x,y) =  \langle f^g_{l,4}(x,y), f^g_{r,4}(x - z,y) \rangle,\\
  & Corr \in \mathbb{R}^{8\times\frac{W}{4}\times\frac{H}{4}\times\frac{W}{4}}, 
\end{aligned}
\end{equation}
where $\langle \cdot,\cdot \rangle$ denotes the inner product, and $g$ represents the group index of the image features. After applying patch embedding and position encoding, the GAPC sequence can be obtained.\par

To reduce memory consumption at large disparities, we adopt the latent space approach proposed in~\cite{jaegle2021perceiver} to compress the disparity dimension into a hidden space. Specifically, a set of learnable $L \times C$ latent tokens is leveraged to perform cross-attention with the GAPC sequence, compressing the variable-length GAPC sequence of shape $(H\times W)\times W \times C$ into a fixed-length latent GAPC sequence $S_0$ with the shape $(H\times W)\times L \times C$.\par

After the sequence is obtained, multiple Bidirectional Attention Blocks are applied, as illustrated in \myfigref{fig:fig_3}(b) (using three in our network), to further extract global geometric and contextual information from GAPC. Specifically, for each block, the latent GAPC sequence is reshaped into $L$ tokens of dimension $C$ per pixel. Self-attention is then applied among these $L$ tokens, followed by a feed-forward network (FFN). This attention mechanism in the latent disparity dimension effectively captures global matching probability distributions for each pixel:
\begin{equation}
\begin{aligned}
   &S_{i} = \text{FFN}(\text{Self-Attention}(S_{i}(1),...,S_{i}(L))),\\
   &\text{for all } i \in H\times W \text{ pixels}.
\end{aligned}
\end{equation}

Then, we transform the latent GAPC sequence into $H\times W$ tokens of dimension $C$. To enrich the geometric and contextual information, the left image features are concatenated along the channel dimension before applying attention. Furthermore, to reduce memory consumption and enhance the capability of capturing both global and local information, we employ Locally-grouped Self-Attention (LSA) and Global Sub-sample Attention (GSA), as proposed in Twins~\cite{chu2021twins}, to perform self-attention on the $H\times W$ tokens:
\begin{equation}
\begin{aligned}
   &S_{i} = \text{LSA}([S_{i}(1),f_l(1)],...,[S_{i}(H\times W),f_l(H\times W)]),\\
   &S_{i} = \text{GSA}([S_{i}(1),f_l(1)],...,[S_{i}(H\times W),f_l(H\times W)]),\\
   &\text{for all } i =1,2,...,L.
\end{aligned}
\end{equation}
\par

After the Bidirectional Attention Blocks, the output is fused with the original latent GAPC sequence via skip connections and element-wise addition and finally transformed into the Global Cost Volume (GCV) with dimensions $C \times L \times \frac{H}{4} \times \frac{W}{4}$. The resulting GCV not only captures the global matching information but also integrates global geometric and contextual details, providing richer information for the subsequent iterative optimization module.\par

\subsection{Local-Global Guided Recurrent Unit-based Iterative Optimization}
\revisethree{During the disparity decoding stage, current ViT-based architectures predominantly inherit the conventional paradigm of coupling a local lookuped cost volume with a ConvGRU-based decoder. This strategy yields the significant benefit of achieving high local precision in disparity predictions. However, a critical limitation of this localized decoding mechanism is the absence of explicit global matching guidance. Consequently, the iterative optimization process often requires an excessive number of iterations to converge and is highly prone to falling into local optima, particularly in ambiguous regions. To overcome this, our fundamental idea is to utilize global contextual information to explicitly guide the fine-grained local matching process. Building upon our proposed LGCV and leveraging cross attention mechanisms, we design a novel Local-Global Guided Recurrent Unit (LGRU) within an iterative optimization framework. This architecture enables the global matching priors to effectively steer the local updates, facilitating faster convergence and yielding highly robust disparity estimates.}

At every iteration, the current disparity $d_{k-1}$ is first utilized to perform a lookup operation on the LCV, obtaining the Lookuped-LCV $L_k$. This tensor is indexed via linear interpolation and has a shape of $C \times R\times \frac{H}{4} \times \frac{W}{4} $, where $R$ denotes the lookup range. Then, we apply a cross-attention mechanism between $L_k$ and the GCV, followed by a feed-forward network (FFN), to extract the globally enhanced matching information $E_k$ for each pixel. Subsequently, the enhanced feature $E_k$, the original $L_k$, and the current disparity $d_{k-1}$ are fed into the motion encoder to generate the motion feature $m_k$, which integrates local matching information, global matching information, and current disparity information. The entire process is formulated as follows:

\begin{equation}
\begin{aligned}
  &L_k = \text{Lookup}(\text{LCV}, d_{k-1})\\
  &E_k = \text{FFN}(\text{Cross-Attention}(L_k, \text{GCV}))\\
  &m_k = \text{Motion Encoder}([E_k, L_k, d_{k-1}])
\end{aligned}
\end{equation}

Next, we employ a multi-level ConvGRU block to update the hidden state $H_{k-1}$, following the approach of Selective-Stereo~\cite{wang2024selective}. The update process is defined as:

\begin{equation}
\begin{aligned}
  &z_k^{s/l} = \sigma( \text{Conv}([H_{k-1}, m_k], W_z^{s/l}) + c_k )\\
  &r_k^{s/l} = \sigma( \text{Conv}([H_{k-1}, m_k], W_r^{s/l}) + c_r )\\
  &\tilde{h}_k^{s/l} = \tanh( \text{Conv}([r_k^{s/l} \odot H_{k-1}, x_k], W_h^{s/l}) + c_h )\\
  &h_k^{s/l} = (1 - z_k^{s/l}) \odot h_{k-1}^{s/l} + z_k^{s/l} \odot \tilde{h}_k^{s/l}\\
  &H_k = A \odot h_k^s + (1-A) \odot h_k^l
\end{aligned}
\end{equation}

where $c_k$, $c_r$, and $c_h$ are context features generated from MGFN. Here, $h_k^s$ represents the GRU with a smaller kernel size, while $h_k^l$ corresponds to the GRU with a larger kernel size. 

Once the hidden state $H_k$ is updated, we utilize a disparity decoder head to predict a residual disparity $\Delta d_k$. The current disparity is then refined as:

\begin{equation}
  d_k = d_{k-1} + \Delta d_k
\end{equation}

\subsection{Loss Function}

We employ two types of loss functions to supervise model training: the initial disparity loss $\mathcal{L}_{init}$ and the iterative disparity loss $\mathcal{L}_{iter}$.\par

The initial disparity loss $\mathcal{L}_{init}$ is defined as:  
\begin{equation}
  \mathcal{L}_{init} = \text{Smooth}_{L_1}(d_0 - d_{gt}),
\end{equation}
where $d_0$ represents the regressed initial disparity map, and $d_{gt}$ denotes the ground-truth disparity map.\par

The iterative disparity loss $\mathcal{L}_{iter}$ is formulated as:  
\begin{equation}
  \mathcal{L}_{iter} = \sum_{n=1}^{N} \gamma^{N-n} \| d_n - d_{gt} \|,
\end{equation}
where $d_n$ ($n \in [1, N]$) represents the disparity map obtained after $n$ iterations, $d_{gt}$ is the ground-truth disparity map, and $\gamma$ is a weighting factor set to 0.9.\par

Finally, the total loss $\mathcal{L}_{total}$ is computed as:  
\begin{equation}
  \mathcal{L}_{total} = \mathcal{L}_{init} + \mathcal{L}_{iter}.
\end{equation}


\section{Experiments}
\subsection{Datasets and Evaluation Metrics}
We pre-train our models using the SceneFlow~\cite{mayer2016large} and Virtual KITTI 2~\cite{cabon2020vkitti2,gaidon2016virtual} datasets and evaluate our model performance on Middlebury~\cite{middlebury}, KITTI 2012~\cite{kitti2012} and KITTI 2015~\cite{kitti2015} 
benchmarks.\par
\textbf{Scene Flow}~\cite{mayer2016large} is a synthetic dataset containing 35,454 training pairs and 4,370 testing frames with dense disparity maps, and all of them are available in cleanpass and finalpass versions, with finalpass being used in training and testing because it is closer to the real-world images.

\par
\textbf{Virtual KITTI 2}~\cite{cabon2020vkitti2, gaidon2016virtual}  is a synthetic dataset of 21,260 image pairs of outdoor autonomous driving scenarios in 5 scenarios. We choose to use it for pre-training because it is more realistic and can help fine-tune the backbone network for better performance.\par
\textbf{KITTI 2012}~\cite{kitti2012} and \textbf{KITTI 2015}~\cite{kitti2015} are datasets collected in real-world driving scenarios and provide sparse \revisetypo{ground} truth collected from LiDAR. The KITTI 2012 dataset consists of 194 training pairs and 195 testing pairs, while the KITTI 2015 dataset includes 200 training pairs and 200 testing pairs. For KITTI 2012, we evaluate performance by reporting the percentage of pixels with errors exceeding \(x\) disparities in both non-occluded regions (\(x\)-noc) and all regions (\(x\)-all). For KITTI 2015, we report the percentage of pixels with an EPE larger than 3 pixels and exceeding 5\% of the true disparity in background regions (D1-bg), foreground regions (D1-fg), and all regions (D1-all).\par
\textbf{Middlebury}~\cite{middlebury} provides a training set with images of 23 indoor scenes and a testing set with images of 10 indoor scenes, and both sets have three resolutions to use and some scenes have different illuminations and exposures. The evaluation metrics include Bad x, which measures the percentage of pixels with an EPE larger than x pixels, and the end-point error (EPE).\par
\subsection{Implementation Details}
We implement our model with the Pytorch framework and perform our experiments using NVIDIA A100 GPUs while using the AdamW optimizer. For pre-training, we trained our model on the augmented Scene Flow~\cite{mayer2016large} training set and Virtual KITTI 2~\cite{cabon2020vkitti2, gaidon2016virtual} for 200k steps with a batch size of 8, and we use a random crop size of 320 $\times$ 736. We use a one-cycle learning rate schedule with a warm-up strategy, and the learning rate gradually increases to 0.0002 in the first 1\% of steps and gradually decreases thereafter. 
\subsection{Comparison on Benchmarks}
\reviseone{We comprehensively evaluate our proposed MLG-Stereo on three standard benchmarks: Middlebury, KITTI 2012, and KITTI 2015.
Notably, at the time of submission, our method achieves highly competitive results across all these public leaderboards, demonstrating the robust generalization capabilities of our architecture.}

\begin{table}[t]
    \centering
    \tabcolsep=3pt
    \caption{Quantitative evaluation on Middlebury\cite{middlebury}. \revisetwo{MLG-Stereo-L and -G represent using DINOv2-L and DINOv2-G~\cite{oquab2023dinov2} as the backbone, respectively.} $^\dagger$As Middlebury only allows a single test submission, we only submitted the results of the G-size model. \textbf{Bold}: Best. \underline{Underline}: Second best. '-' indicates that the corresponding data is not available in both the original paper and the benchmark leaderboard.}
    \setlength{\tabcolsep}{1.6mm}
    \begin{tabular}{l|c c c}
     \toprule
     Method & bad1.0 & bad2.0 & EPE  \\
     \midrule
    PSMNet(CVPR 2018)~\cite{psmnet} & 63.9 & 42.1 & 6.68 \\
    ADStereo(TIP 2025)~\cite{10890914} & 35.7 & 18.0 & 4.78 \\
    RAFT-Stereo(3DV 2021)~\cite{raft-stereo} & 9.37 & 4.74 & 1.27 \\
    CREStereo(CVPR 2022)~\cite{crestereo} & 8.25 & 3.71 & 1.15 \\
    GMStereo(TPAMI 2023)~\cite{xu2023unifying} & 23.6 & 7.14 & 1.31 \\
    IGEV-Stereo(CVPR 2023)~\cite{igevstereo} & 9.41 & 4.83 & 2.89 \\
    DLNR(CVPR 2023)~\cite{zhao2023high} & 6.82 & 3.20 & 1.06 \\
    Selective-IGEV(CVPR 2024)~\cite{wang2024selective} & 6.53 & 2.51 & 0.91 \\
    AIO-Stereo(AAAI 2025)~\cite{zhou2024all} & 6.08 & 2.36 & 0.85 \\
    \reviseone{FoundationStereo(CVPR 2025)~\cite{wen2025foundationstereo}} & \underline{4.39} & \underline{1.84} & \underline{0.78} \\
    \reviseone{CST-Stereo(CVPR 2025)~\cite{zhou2025consistency}} & 6.23 & 2.40 & 0.87 \\
    \reviseone{DEFOM-Stereo(CVPR 2025)~\cite{jiang2025defom}} & 5.81 & 2.39 & 0.79 \\
    \reviseone{Monster(CVPR 2025)~\cite{cheng2025monster}} & - & 2.64 & - \\
    \reviseone{IGEV++(TPAMI 2025)~\cite{xu2025igev++}} & 7.18 & 3.23 & 0.97 \\
    \midrule
    MLG-Stereo-G(Ours)$^\dagger$ & \textbf{4.30} & \textbf{1.76} & \textbf{0.74} \\
    \bottomrule
    \end{tabular}
    \label{tab:sota_middlebury}
\end{table}

\begin{table*}[t]
    \centering
    \tabcolsep=3pt
    \caption{Quantitative evaluation on KITTI 2012~\cite{kitti2012} and KITTI 2015~\cite{kitti2015}. \revisetwo{MLG-Stereo-L and -G represent using DINOv2-L and DINOv2-G~\cite{oquab2023dinov2} as the backbone, respectively.} * means using Depth Anything 2~\cite{depth_anything_v2} as backbone. $^\ddagger$ indicates the results are based on the CF-Net~\cite{shen2021cfnet} version. \textbf{Bold}: Best. \underline{Underline}: Second best. '-' indicates that the corresponding data is not available in both the original paper and the benchmark leaderboard.}
    \setlength{\tabcolsep}{1.6mm}
    \begin{tabular}{l|ccc|c c c c}
     \toprule
     \multirow{2}{*}{Method} & \multicolumn{3}{c|}{ KITTI 2015 } & \multicolumn{4}{c}{ KITTI 2012 } \\
     \cmidrule(lr){2-4}\cmidrule(lr){5-8}
      & D1-bg & D1-fg & D1-all & 2-noc & 2-all & 3-noc & 3-all \\
     \midrule
    PSMNet(CVPR 2018)~\cite{psmnet} & 1.86 & 4.62 & 2.32 & 2.44 & 3.01 & 1.49 & 1.89 \\
    ADStereo(TIP 2025)~\cite{10890914} & 1.59 & 2.94 & 1.82 & 2.23 & 2.69 & 1.36 & 1.68 \\
    RAFT-Stereo(3DV 2021)~\cite{raft-stereo} & 1.75 & 2.89 & 1.96 & 1.92 & 2.42 & 1.30 & 1.66 \\
    CFNet-RSSM(TMM 2022)~\cite{cheng2022region}&1.46&3.00&1.72&-&-&-&- \\
    CREStereo(CVPR 2022)~\cite{crestereo} & 1.45 & 2.86 & 1.69 & 1.72 & 2.18 & 1.14 & 1.46 \\
    GMStereo(TPAMI 2023)~\cite{xu2023unifying} & 1.49 & 3.14 & 1.77 & - & - & - & - \\
    VITAStereo(TIV 2024)~\cite{li2024roadformer} & 1.21 & 2.99 & 1.50 & 1.46 & 1.80 & 0.93 & 1.16 \\
    IGEV-Stereo(CVPR 2023)~\cite{igevstereo} & 1.38 & 2.67 & 1.59 & 1.71 & 2.17 & 1.12 & 1.44 \\
    DLNR(CVPR 2023)~\cite{zhao2023high} & 1.60 & 2.59 & 1.76 & - & - & - & - \\
    UPFNet(TCSVT 2023)~\cite{chen2023unambiguous}&1.26&2.70&1.50&1.79&2.28&1.12&1.46\\
    DMCA-Net(TCSVT 2024)~\cite{zeng2023deep}&1.45&2.78&1.67&1.83&2.35&1.13&1.48\\
    Selective-IGEV(CVPR 2024)~\cite{wang2024selective} & 1.33 & 2.61 & 1.55 & 1.59 & 2.05 & 1.07 & 1.38 \\
    Reg-Stereo(TCSVT 2025)~\cite{zhu2025region}&1.30&2.54&1.50&-&-&1.04&1.35\\
    AIO-Stereo(AAAI 2025)~\cite{zhou2024all} & 1.35 & 2.46 & 1.54 & 1.58 & 1.94 & 1.05 & 1.29 \\
    \reviseone{CST-Stereo(CVPR 2025)~\cite{zhou2025consistency}} & 1.30 & 2.48 & 1.50 & - & - & - & - \\
    \reviseone{DEFOM-Stereo(CVPR 2025)~\cite{jiang2025defom}} & 1.25 & \underline{2.23} & 1.41 & 1.43 & 1.79 & 0.94 & 1.18 \\
    \reviseone{Monster(CVPR 2025)~\cite{cheng2025monster}} & \textbf{1.13} & 2.81 & 1.41 & \textbf{1.36} & 1.75 & \textbf{0.84} & \underline{1.09} \\
    \reviseone{IGEV++(TPAMI 2025)~\cite{xu2025igev++}} & 1.31 & 2.54 & 1.51 & - & - & - & - \\
    \reviseone{IGEV++*(TPAMI 2025)~\cite{xu2025igev++}} & \underline{1.15} & 2.80 & 1.43 & \textbf{1.36} & \underline{1.74} & 0.89 & 1.13 \\


    \midrule
    MLG-Stereo-L(Ours) & 1.23 & \underline{2.23} & \underline{1.40} & 1.41 & 1.76 & 0.91 & 1.13 \\
    MLG-Stereo-G(Ours) & 1.19 & \textbf{2.17} & \textbf{1.36} & 1.37 & \textbf{1.71} & \underline{0.86} & \textbf{1.08} \\
    \bottomrule
    \end{tabular}
    \label{tab:sota_kitti}
\end{table*}

\subsubsection{Middlebury}
 \begin{figure*}[t]
  \centering
   \includegraphics[width=1.0\linewidth]{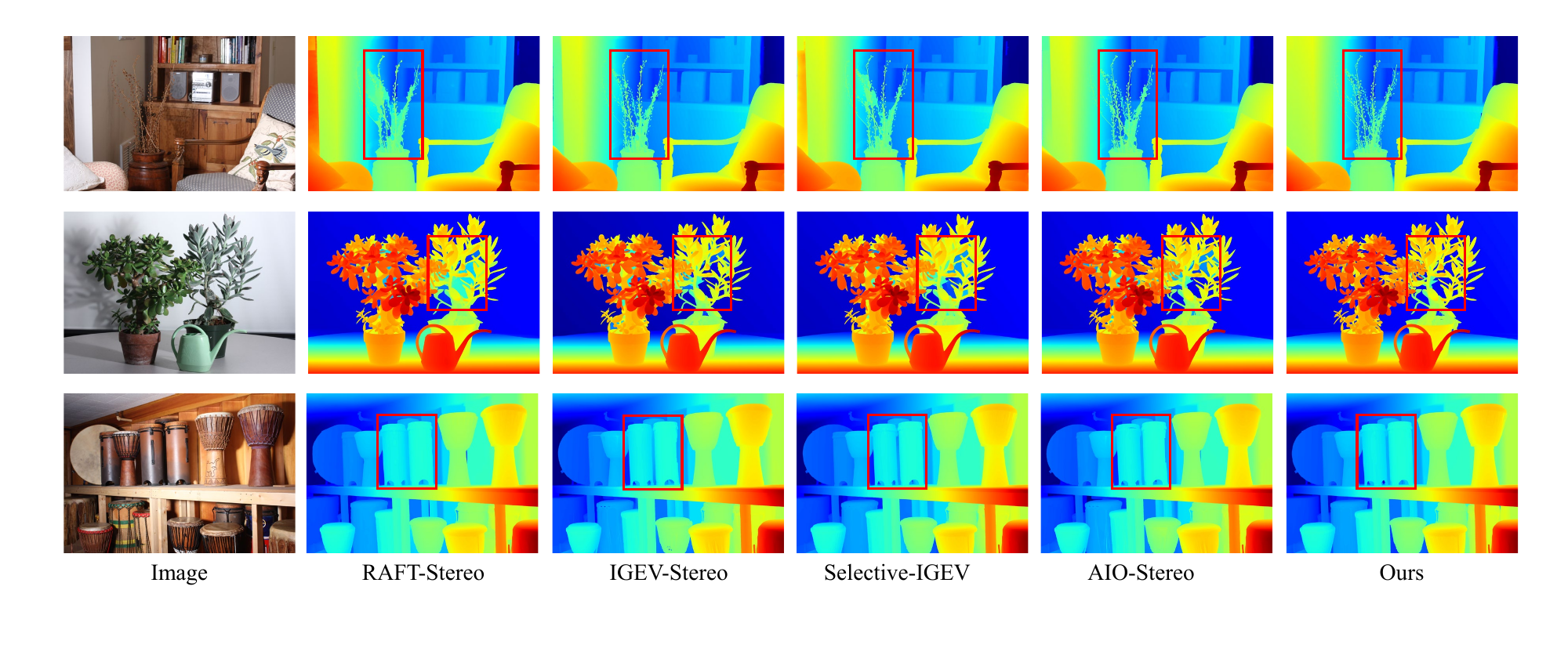}
   \caption{Visual comparison on the MiddEval v3 test set. Our model not only can infer ultra-high resolution images like CNN models, but also has better performance in difficult areas. }
   \label{fig:fig_7}
\end{figure*}
For the Middlebury dataset, following~\cite{wang2024selective}, we first finetune our pre-trained model on the mixed Tartan Air~\cite{wang2020tartanair}, CREStereo Dataset~\cite{crestereo}, Scene Flow, Falling things~\cite{tremblay2018falling}, InStereo2k~\cite{bao2020instereo2k}, CARLA HR-VS~\cite{yang2019hierarchical}, and Middlebury datasets 200k steps using a crop size of 384 $\times$ 512 with a batch size of 8. Then we finetune it on the mixed CREStereo Dataset, Falling Things, InStereo2k, CARLA HR-VS, and Middlebury datasets using a crop size of 384 $\times$ 768 with a batch size of 8 for another 100k steps. \reviseone{As shown in Table~\ref{tab:sota_middlebury}, our method achieves highly competitive performance on the Middlebury test set. Specifically, our method outperforms the recent leading methods AIO-Stereo~\cite{zhou2024all} and FoundationStereo~\cite{wen2025foundationstereo} by 25.42\% and 4.35\% on the Bad 2.0 metric. Partial prediction visualization is shown in \myfigref{fig:fig_7}. Thanks to our multi-stage local global enhancement approach, our model robustly adapts to images with completely different resolution scales from the pre-training set while maintaining precise edge predictions.}
\subsubsection{KITTI}

 \begin{figure*}[t]
  \centering
   \includegraphics[width=1.0\linewidth]{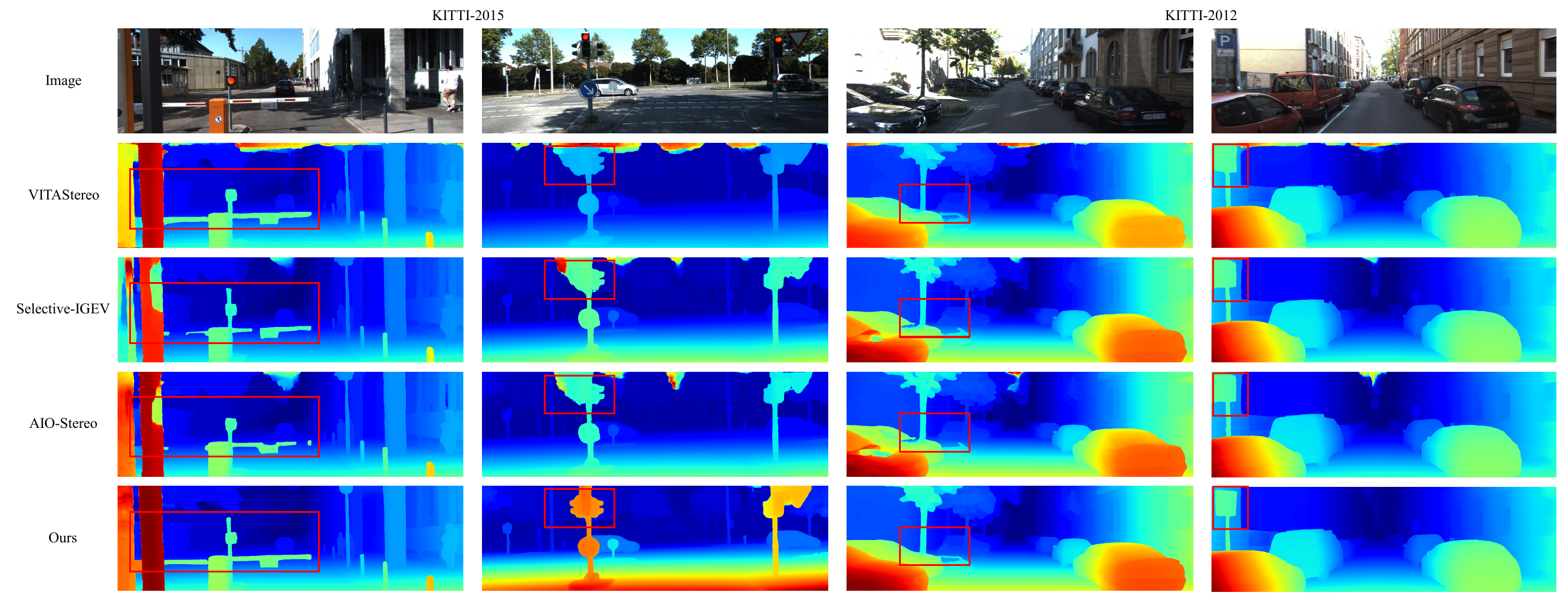}
   \caption{Visual comparison on the KITTI-2015 and KITTI-2012 test set. Compared to CNN-based models, our model has \revisetypo{a stronger predictive ability in challenging regions} (such as weak textures, hollows, etc.); Compared to other ViT-based models, our model has \revisetypo{a stronger capability for detail prediction}.}
   \label{fig:fig_6}
\end{figure*}
For the KITTI dataset, following ~\cite{wang2024selective}, we finetune the pretrained model on the mixed dataset of KITTI-2012~\cite{kitti2012} and KITTI-2015~\cite{kitti2015} with a batch size of 8 for 50k steps. \reviseone{It is worth noting that due to the relatively low resolution and sparse ground truth of the KITTI dataset, the performance of current mainstream models has largely saturated, with residual errors primarily concentrated in a minimal fraction of extremely challenging regions. Despite this saturated context, as shown in Table~\ref{tab:sota_kitti}, at the time of submission, our method still achieves the lowest D1-all error on KITTI-2015. Specifically, our method outperforms the leading methods AIO-Stereo~\cite{zhou2024all} and Monster~\cite{cheng2025monster} by 11.69\% and 3.55\% on D1-all metric. Moreover, as shown in \myfigref{fig:fig_6}, thanks to our multi-stage local global enhancement approach, our model demonstrates stronger robustness precisely in these remaining difficult regions (e.g., hollows or detail areas).}
\subsection{Ablation Study}
For ablation experiments, we pretrain our model \revisetypo{for} 100k rounds in the case that the dataset and other hyperparameters are the same. \revisetypo{Then,} we finetune the model on the Mixed Middlebury dataset for 6k steps while the learning rate linearly decays from $3e^{-5}$ to 0. In order to compare the performance of the models at different resolutions more easily and fairly, we used three different \revisetypo{resolution training sets} of the MiddEval v3 benchmark \revisetypo{that were excluded during training} to evaluate the models.

\subsubsection{Module Ablation}
\revisethree{To verify the effectiveness of each proposed component in our framework, we conduct a comprehensive ablation study on the Middlebury dataset. As presented in Table~\ref{tab:module_ablation}, the baseline model (Selective-IGEV~\cite{wang2024selective}) exhibits limited performance across all resolution settings. By integrating the MGFN, the bad 2.0 error is reduced by 19.97\%, 27.37\% and 48.36\% on Full, Half and Quarter resolution, respectively. This significant improvement demonstrates that our MGFN effectively captures multi-scale features, enhancing robustness against resolution variations. Furthermore, the introduction of the LGCV and LGRU brings additional performance gains. Specifically, adding LGCV alone further reduces the error on full-resolution images, indicating its capability in handling large disparities through global attention. Similarly, incorporating LGRU improves the optimization process, leading to better convergence. Finally, the full model (MGFN + LGCV + LGRU) achieves the best overall performance, validating the complementary nature of our proposed modules in balancing global context and local details for accurate stereo matching.}
\begin{table}[t]
  \centering
  \tabcolsep=3pt
  \caption{Ablation study of different components on Middlebury dataset. The Baseline is Selective-IGEV~\cite{wang2024selective}.}
    \begin{tabular}{l|cc|cc|cc}
    \toprule
     \multirow{2}{*}{Method}   &\multicolumn{2}{c|}{Middlebury F} & \multicolumn{2}{c|}{Middlebury H}& \multicolumn{2}{c}{Middlebury Q}\\
     \cmidrule(lr){2-3}\cmidrule(lr){4-5}\cmidrule(lr){6-7}
      &  EPE&Bad2.0&  EPE&Bad2.0&  EPE&Bad2.0  \\
    \midrule
    Baseline &1.31 & 7.11 & 0.79 & 4.86 & 0.83 & 7.01 \\
    Baseline+MGFN & 1.21 & 5.69 & 0.65 & 3.53 & 0.60 & 3.62 \\
    Baseline+MGFN+LGCV &0.91&4.82&0.64&3.45&0.65&4.76\\
    Baseline+MGFN+LGRU & \textbf{0.88} & 4.74 & 0.67 & 3.65 & 0.63 & 4.68 \\
    Full & 0.89 & \textbf{4.53} & \textbf{0.62} & \textbf{3.33} & \textbf{0.58} & \textbf{3.34} \\
    \bottomrule
  \end{tabular}
  \label{tab:module_ablation}
\end{table}

\begin{figure*}[t]
  \centering
   \includegraphics[width=1\linewidth]{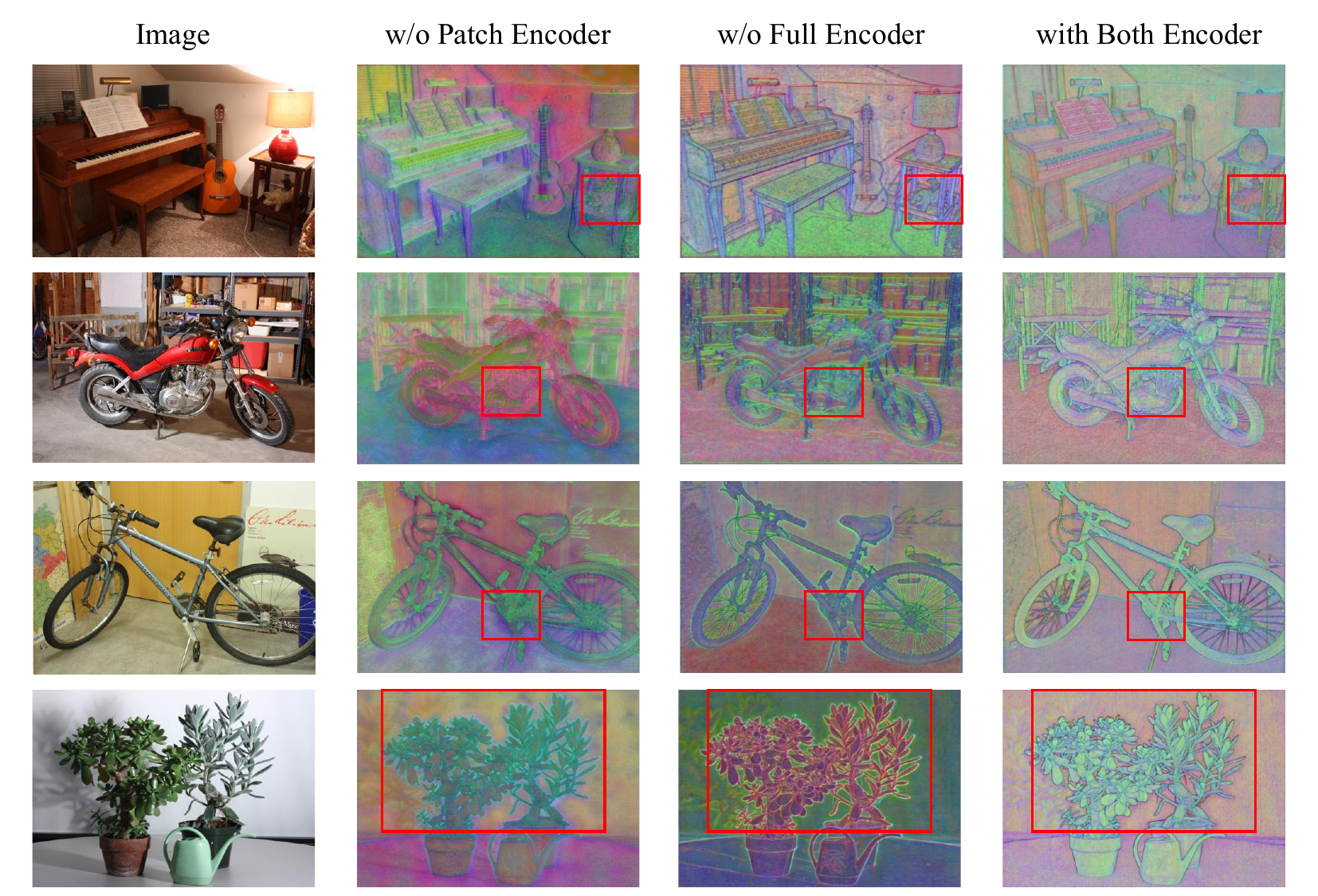}
   \caption{Visualization of image features extracted by encoder after PCA, \revisetypo{from} left to right: without patch image encoder, without full image encoder, with both image encoder. The features extracted using our proposed MGFN can simultaneously consider both local geometric features and global semantic features. }
   \label{fig:fig_4}
\end{figure*}
\subsubsection{Effectiveness of MFGN}
\begin{table}[t]
  \centering
  \tabcolsep=3pt
  \caption{
  Ablation study of the Encoder part. The baseline is the \revisetypo{Selective-IGEV}~\cite{wang2024selective}, while all \revisetypo{other methods use the same network structure except for the encoder}. * means the final version of our method. }
    \begin{tabular}{l|cc|cc|cc}
    \toprule
     \multirow{2}{*}{Method}   &\multicolumn{2}{c|}{Middlebury F} & \multicolumn{2}{c|}{Middlebury H}& \multicolumn{2}{c}{Middlebury Q}\\
     \cmidrule(lr){2-3}\cmidrule(lr){4-5}\cmidrule(lr){6-7}
      &  EPE&Bad2.0&  EPE&Bad2.0&  EPE&Bad2.0  \\
    \midrule
    Baseline &1.31 &7.11 &0.79 &4.86 &0.83 &7.01 \\
    \midrule
    w/o MGFN& 1.19	&6.72	&0.89	&6.58	&0.83&	6.31\\
    w/o patch image encoder  & 1.70 &8.27 &0.78 &5.51 &0.71 &5.81 \\
    \reviseone{w/ CNN patch encoder} & \reviseone{2.01} & \reviseone{7.64} & \reviseone{0.78} & \reviseone{5.24} & \reviseone{0.71} & \reviseone{5.13} \\
    w/o full image \revisetypo{encoder} &1.07 &6.63 &0.83 &6.53 &0.81 &7.02 \\
    w/o fusion network&1.03 &6.35 &0.72 &4.74 &0.72 &5.68 \\
    w/o half size patch & 2.02&6.09 &0.69 &3.67 &0.66 &4.8 \\
    \midrule
    Full*   &\textbf{0.89}& \textbf{4.53}& \textbf{0.62}& \textbf{3.33}&\textbf{0.58} &\textbf{3.34}  \\
    \bottomrule
  \end{tabular}
  
  \label{tab:ablation1}
\end{table}
In order to verify the effectiveness of MGFN which can extract high-quality image features at different resolutions, \revisetypo{we conducted} the ablation experiments as shown in \revisetypo{Table}~\ref{tab:ablation1}. First, we remove the MGFN from \revisetypo{the} full model, which showed a significant decrease compared to the full model at all three resolutions. This proves that the extracted features of the proposed MGFN are of higher quality than the baseline model's. Subsequently, we studied the influence of the MGFN structure \revisetypo{on} the results, for which we removed the full-image encoder, patch image encoder, half resolution patch image input and fusion network, respectively. \reviseone{We also replaced the ViT-based patch image encoder with a CNN-based encoder to verify the necessity of using ViT for extracting local features.} The results show that the performance of the model decreases not only relative to the final model, but also decreases in some indicators even compared with the baseline model(\emph{i.e.}, without \revisetypo{the} patch image encoder, the model is weaker than the baseline model in high-resolution image prediction). In addition, we also perform principal component analysis (PCA) on the features extracted by the encoder. As shown in~\myfigref{fig:fig_4}, removing the patch image encoder leads to the loss of fine details, while discarding the full image encoder results in a degradation of global semantic information. Our method combines the strengths of both, effectively balancing local details and global context. Finally, with the help of \revisetypo{the} feature fusion networks, high-quality image features that contain both local detail information and global semantic information can be obtained. This shows that the structure of the MGFN we designed can robustly extract high-quality image features at different resolutions.
\subsubsection{Design of LGCV}
\begin{table}[t]
  \centering
  \tabcolsep=2pt
  \caption{
  Ablation study of the Decoder part. The baseline is the \revisetypo{Selective-IGEV}~\cite{wang2024selective}, while all \revisetypo{other methods share} the same network structure apart from \revisetypo{the} Decoder. * means the final version of our method. }
    \begin{tabular}{l|cc|cc|cc}
    
    \toprule
    \multirow{2}{*}{Method}   &\multicolumn{2}{c|}{Middlebury F} & \multicolumn{2}{c|}{Middlebury H}& \multicolumn{2}{c}{Middlebury Q}\\
     \cmidrule(lr){2-3}\cmidrule(lr){4-5}\cmidrule(lr){6-7}
      &  EPE&Bad2.0&EPE&Bad2.0&EPE&Bad2.0\\
    \midrule
    Baseline &1.31 &7.11 &0.79 &4.86 &0.83 &7.01 \\
    \midrule
    w/o Bidirectional Attention&1.04&4.92&0.64&\textbf{3.27}&0.61&4.03\\
    w/o Disparity Attention&0.92 &4.75 &0.64 &3.33 &0.63 &4.41 \\
    w/o \revisetypo{Spatial} Attention &1.9 &5.53 &0.65 &3.65 &0.66 &4.63 \\
    \midrule
    \revisethree{w/o LGRU} &0.91&4.82&0.64&3.45&0.65&4.76\\
    w/o Lookuped-LCV&2.03 &12.19 &1.01 &8.93 &1.06 &10.15 \\
    \midrule
    Full*   &\textbf{0.89}& \textbf{4.53}& \textbf{0.62}& 3.33&\textbf{0.58} &\textbf{3.34}  \\
    \bottomrule
  \end{tabular}
  
  \label{tab:ablation2}
\end{table}
To explore the design of LGCV, we adopted various model structures as shown in \revisetypo{Table}~\ref{tab:ablation2}. Firstly, we directly input GAPC into the subsequent iterative decoding block, and the results show that the performance \revisetypo{at medium/high resolutions} has decreased. Subsequently, we only used Spatial Attention and Disparity Attention separately. As shown in \revisetypo{Table}~\ref{tab:ablation2}, without Spatial Attention, due to the lack of additional image \revisetypo{feature input}, the performance of \revisetypo{the model significantly decreases} at high resolutions, while without Disparity Attention, the performance of the model \revisetypo{decreases} at low/medium resolutions. 
\subsubsection{Ablation of LGRU}

\begin{figure}[t]
  \centering
   \includegraphics[width=1.0\linewidth]{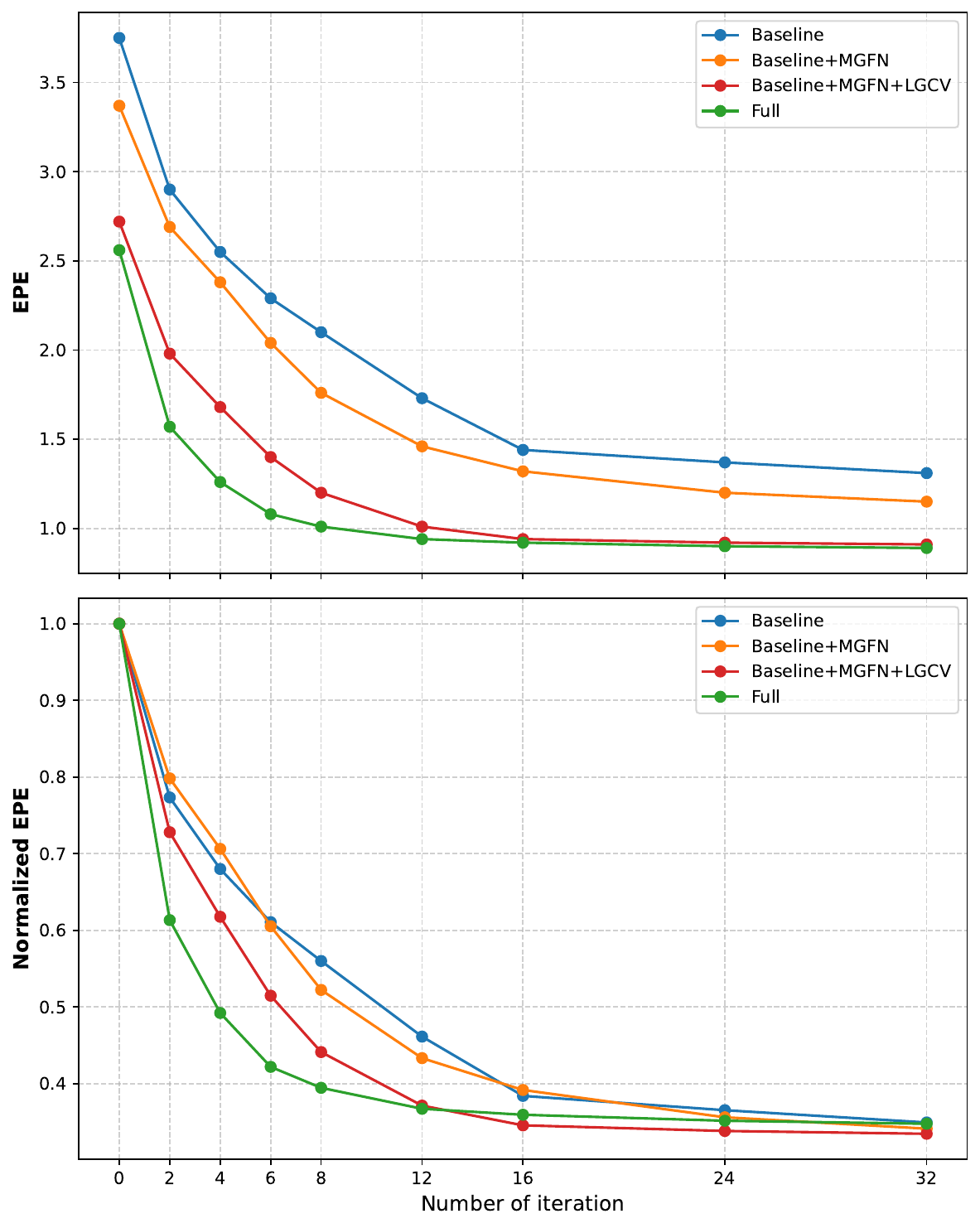}
   \caption{The variation of model performance on the Middlebury dataset with \revisetypo{the number of iterations}. Our model not only outperforms the baseline model \revisetypo{with the same number of iterations}, but also converges in fewer iterations}
   \label{fig:fig_5}
\end{figure}
To verify the effectiveness of our LGRU design, we first removed the cross-attention block and instead directly sampled the LCV and GCV locally. The results show that the performance decreases to varying degrees across all resolutions. Then we remove the concatenation of the Lookuped-LCV after the cross-attention block, and the results show that the prediction performance at all resolutions also decreases. All quantitative results are shown in \revisetypo{Table}~\ref{tab:ablation2}.
\subsubsection{Number of Iteration}
Our MLG-Stereo can achieve better results with fewer iterationss. As shown in~\myfigref{fig:fig_5}, our model outperforms the baseline model Selective-IGEV~\cite{wang2024selective} in all iterations and only requires 4 iterations to achieve the same performance as the baseline model. In addition, our model's results can converge in fewer iterations. This proves that our method of decoding disparity using both global and local matching information can help the model achieve better results faster.

\subsection{Zero-Shot Generalization}
\revisetwo{Following IGEV-Stereo~\cite{igevstereo}, we evaluate the model trained only on the synthetic SceneFlow dataset on the MiddEval v3 training set to test its zero-shot generalization capability.} As shown in \revisetypo{Table}~\ref{tab:zeroshot}, MLG-Stereo achieves state-of-the-art performance \revisetypo{on} most of the metrics. Attributed to the multi-granularity feature network (MGFN) based on the VFM and the matching and iterative decoding of both local and global factors, our model has robust and excellent performance \revisetypo{on} unseen datasets of different resolutions.

 \begin{figure*}[!ht]
  \centering
   \includegraphics[width=1.0\linewidth]{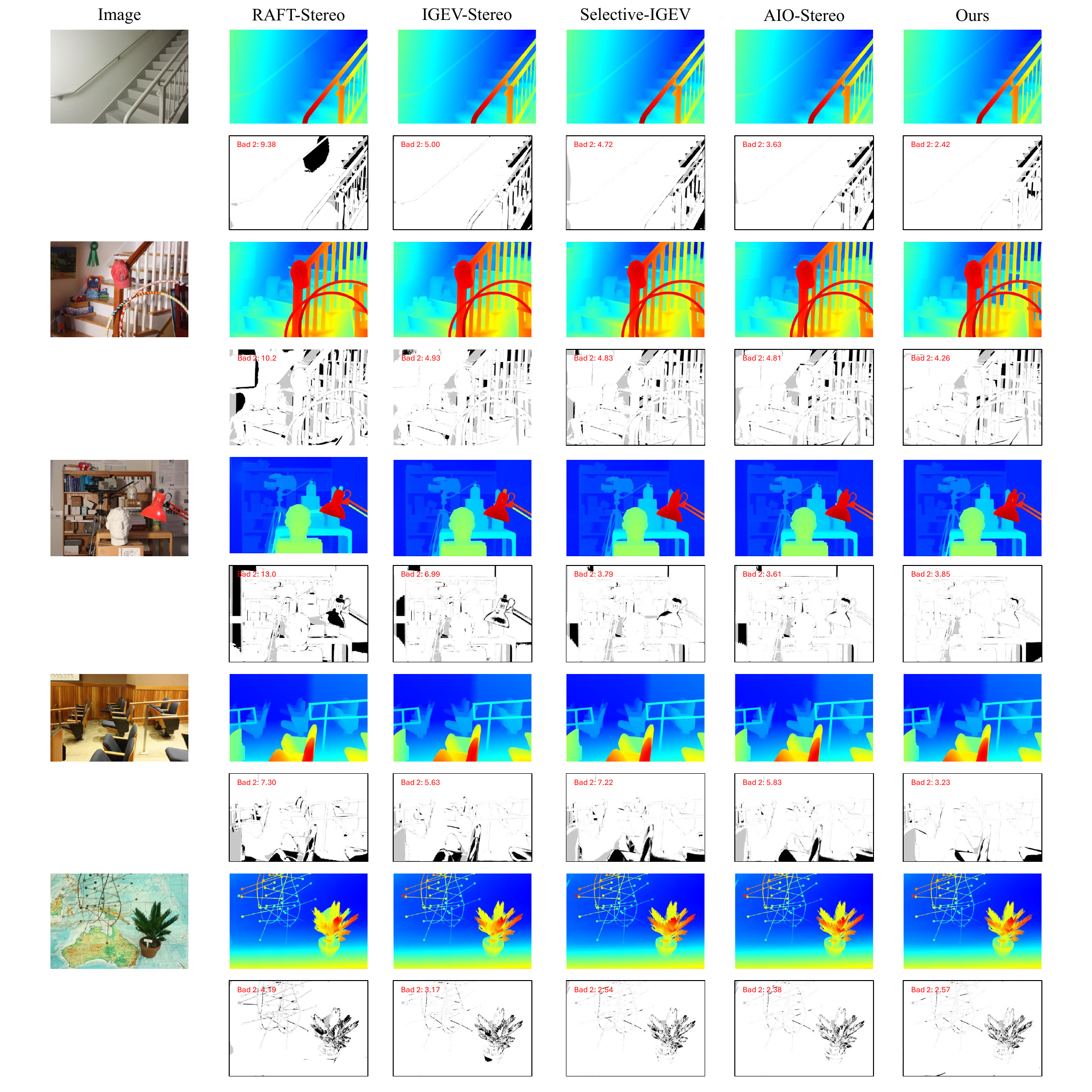}
  \caption{\reviseone{Challenging cases analysis on the Middlebury benchmark. We sort the validation set by per-image error and visualize the worst-performing samples (bottom 5 cases) along with their error masks. In the error mask, black regions represent erroneous pixels in non-occluded areas, while gray regions indicate erroneous pixels in occluded areas. The remaining prediction errors are primarily localized in extreme ill-posed regions, such as complex occlusions and textureless edges.}}
   \label{fig:fig_7}
\end{figure*}

\footnotetext{Since this work did not provide a model trained only on the SceneFlow dataset, we used a community reproduction model available at \url{https://github.com/XiandaGuo/OpenStereo}.}
\section{Limitations and Future Work}

\begin{table*}[!th]
    \centering
    \caption{Computational analysis on Middlebury Benchmark. Time is measured in seconds, Memory using peak VRAM in GB, and computation in TFLOPS.}
    \label{tab:comp_analysis}
    \setlength{\tabcolsep}{4.5mm}{
    \begin{tabular}{l|r r r|r r r}
    \toprule
    Method & Time (s) & Memory (GB) & TFLOPS &Bad 1.0&Bad 2.0&EPE\\
    \midrule
    Selective-IGEV&  1.922 & 10.34 & 33.94&6.53&2.51&0.91 \\
    FoundationStereo& 12.453 & 44.75 & 136.95&4.39&1.84&0.78 \\
    MLG-Stereo-G (Ours)&8.636 & 26.27 & 128.68&\textbf{4.30}&\textbf{1.76}&\textbf{0.74} \\
    \bottomrule
    \end{tabular}
    }
\end{table*}
\begin{table}[!th]
\centering
  \centering
  \caption{Zero-shot evaluation on the Middlebury dataset. \textbf{Bold}: Best. \underline{Underline}: Second best.}
  \tabcolsep=3pt

  \begin{tabular}{l|cc|cc|cc}
  \toprule
  \multirow{2}{*}{Method}   &\multicolumn{2}{c|}{Middlebury F} & \multicolumn{2}{c|}{Middlebury H}& \multicolumn{2}{c}{Middlebury Q}\\
  \cmidrule(lr){2-3}\cmidrule(lr){4-5}\cmidrule(lr){6-7}
  &EPE&Bad2.0&EPE&Bad2.0&EPE&Bad2.0\\
  \midrule
  PSMNet~\cite{psmnet} &40.51&57.93&9.79&32.19&2.29&18.17\\
  RAFT-Stereo~\cite{raft-stereo} &\underline{3.84}&15.64&1.44&11.21&0.93&7.44\\
  IGEV-Stereo~\cite{igevstereo}&5.87&11.85&1.36&7.21&0.76&6.21 \\
  GMStereo~\cite{xu2023unifying}&4.10&29.15&1.92&15.69&1.06&9.12\\
  EAI-Stereo~\cite{zhao2022eai}&6.16&18.25&2.15&11.74&1.13&9.91\\
  DLNR~\cite{zhao2023high} &6.57&14.46&1.45&9.46&0.86&7.63\\
  Selective-IGEV~\cite{wang2024selective} &5.28&12.07&0.98&7.10&0.85&6.57\\
  AIO-Stereo ~\cite{zhou2024all} &4.16&11.67&0.89&6.48&0.79&5.91\\
  \revisetwo{DEFOM-Stereo ~\cite{jiang2025defom}} &\textbf{2.12}&\underline{11.26}&\underline{0.84}&\underline{6.03}&\underline{0.62}&5.53\\
  \revisetwo{FoundationStereo ~\cite{wen2025foundationstereo}\footnotemark} &5.27&15.24&1.04&7.56&0.63&\underline{4.48}\\
  MLG-Stereo(Ours)&5.53&\textbf{9.71}&\textbf{0.76}&\textbf{3.93}&\textbf{0.53}&\textbf{3.63}\\
  \bottomrule
  \end{tabular}
  
  \label{tab:zeroshot}
\end{table}
\reviseone{Despite achieving highly competitive performance on major benchmarks, our proposed MLG-Stereo exhibits certain limitations that we aim to address in future research.}

\subsection{Accuracy--Efficiency Trade-off}

\reviseone{While the Multi-Granularity Feature Network (MGFN) based on the DINOv2 backbone ensures high-quality feature extraction, it inevitably introduces a computational burden compared to traditional CNNs. To ensure a fair comparison, all models are evaluated with 32 iterative updates using half-precision floating-point (BF16) computation, with a maximum disparity set to 768. The quantitative results are reported in Table~\ref{tab:comp_analysis}. Compared to the CNN-based Selective-IGEV, although our MLG-Stereo requires more computational resources (e.g., inference time and memory) due to the heavy computation load inherent in the large-scale VFM backbone, it achieves significantly stronger performance (reducing Bad 2.0 by \textbf{29.9\%}). Moreover, compared to FoundationStereo, another VFM-based method, our model demonstrates clear advantages in both accuracy and efficiency. Notably, thanks to our strategy of compressing disparity into the latent space, our computational cost is significantly lower under ultra-large disparity settings. For example, on Middlebury Full resolution, MLG-Stereo-G not only achieves lower error rates (Bad 2.0: 1.76 vs 1.84) but is also notably more efficient in terms of time and memory (8.636s vs 12.453s; 26.27GB vs 44.75GB).}

\reviseone{Nevertheless, the relatively high parameter count and inference latency make it challenging to deploy the current model on resource-constrained edge devices. To mitigate this, our future work will focus on model lightweighting. We plan to utilize knowledge distillation techniques to guide the training of compact student networks and explore model sparsification strategies to reduce computational redundancy without significant performance degradation.}

\subsection{Challenges in Ill-posed Regions}

\reviseone{To systematically analyze the theoretical boundaries of our method, we evaluate its performance on challenging cases. Specifically, we select the bottom 5 worst-performing samples from the Middlebury benchmark based on the quantitative error rank, as visualized in Fig.~\ref{fig:fig_7}. As shown in the error masks, unresolved challenges are primarily concentrated in extreme ill-posed regions, including: (1) mutually occluded regions with complex structures (e.g., thin railings or hollow areas); and (2) edge and boundary regions adjacent to weak textures. In these localized regions, matching ambiguity remains a common challenge for current stereo matching architectures.}

\reviseone{We attribute these residual errors to the current reliance on the generic DINOv2 backbone, which offers robust semantic visual representations but lacks explicit depth-oriented geometric priors. As evidenced in the ablation studies of FoundationStereo~\cite{wen2025foundationstereo}, incorporating depth-specific foundation models can provide stronger guidance for such localized ambiguities. Therefore, in future iterations, we plan to systematically integrate depth-aware backbones like Depth Anything 3~\cite{lin2025depth}. By exploring these large-scale depth priors, we anticipate further mitigating the matching ambiguity in extreme textureless, multi-occluded, and fine-edge regions.}
\section{Conclusion}

\reviseone{In this paper, we presented MLG-Stereo, a unified framework that couples local-global enhancement across the entire stereo matching pipeline. By integrating the Multi-Granularity Feature Network (MGFN), Local-Global Cost Volume (LGCV), and Local-Global Guided Recurrent Unit (LGRU), our approach demonstrates that local-global complementarity is highly effective in the feature encoder, matching volume, and iterative refinement stages. This structural design accurately recovers high-frequency geometric details while maintaining the robust zero-shot capabilities of Vision Foundation Models.}

\reviseone{Despite achieving highly competitive performance on major benchmarks, MLG-Stereo still exhibits certain performance boundaries. First, the reliance on a full-size DINOv2 backbone inevitably introduces high computational overhead and inference latency. Second, although our method demonstrates strong global matching capabilities, ambiguous areas such as mutually occluded regions and thin structures with weak textures remain challenging, indicating a need for more explicit depth priors.}

\reviseone{To address these challenges, our future work will primarily focus on two directions. First, we plan to implement lightweight strategies, including knowledge distillation and model sparsification, to reduce computational redundancy for practical deployment. Second, we aim to integrate stronger depth-specific priors, such as substituting the backbone with depth-oriented foundation models, to further enhance matching stability and robustness in challenging regions.}




\bibliographystyle{IEEEtran}
\bibliography{references}
\begin{IEEEbiography}[{\includegraphics[width=1in,height=1.25in,clip,keepaspectratio]{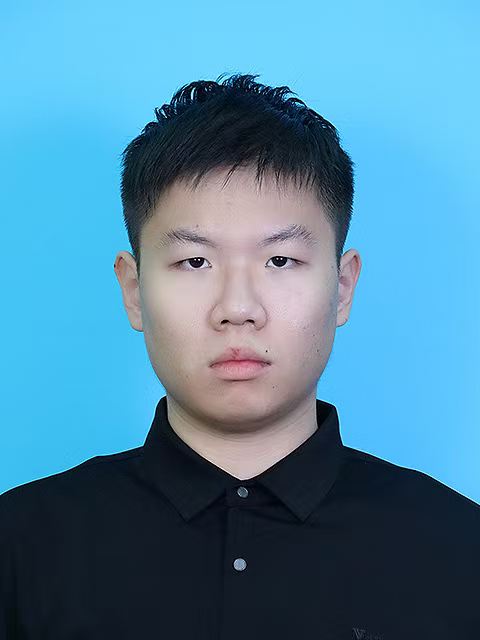}}]{Haoyu Zhang}
received the B.S. degree in electronic science and technology from the School of Information Science and Technology, Fudan University, Shanghai, China, in 2025. 
He is currently pursuing the Ph.D. degree with the College of Future Information Technology, Fudan University.
His research interests include computer vision, multimodal large models, and intelligent agent systems. 
\end{IEEEbiography}

\begin{IEEEbiography}[{\includegraphics[width=1in,height=1.25in,clip,keepaspectratio]{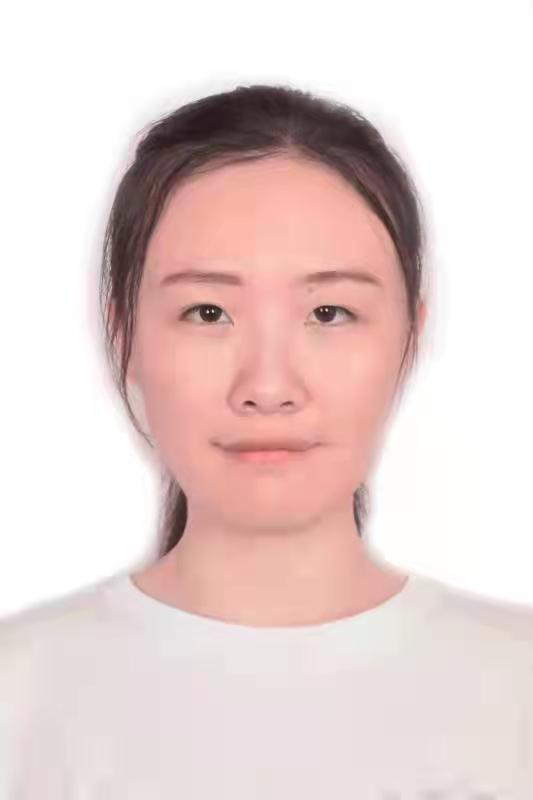}}]{Jingyi Zhou}
received the B.E. degree from the School of Information Science and Technology, Fudan University, Shanghai, China, in 2023, and she is currently pursuing the M.E. degree in the School of Future Information Innovation, Fudan University, Shanghai, China. Her main research interests include computer vision, hyperspectral analysis, sentiment analysis and depth estimation.
\end{IEEEbiography}
\begin{IEEEbiography}[{\includegraphics[width=1in,height=1.25in,clip,keepaspectratio]{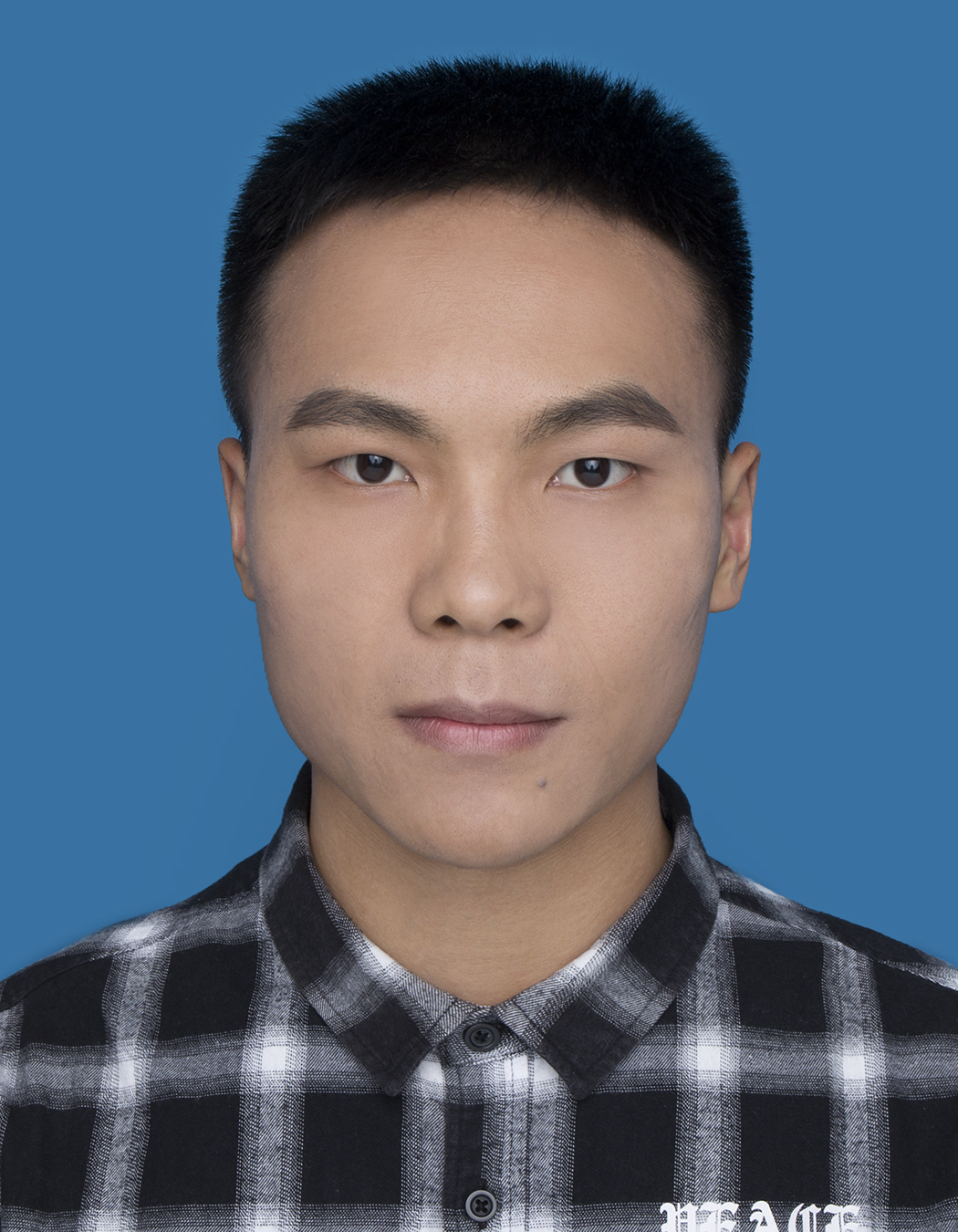}}]{Peng Ye}
is currently a postdoctoral fellow at mmlab of the Chinese University of Hong Kong and a scientific research advisor of Shanghai AI Lab. He received his Ph.D. degree at Fudan University, Shanghai, China. His research interests include Autonomous Agent, (M)LLM, Foundation Model, Efficient Design and Optimization. He has published papers in leading journals and conferences including PAMI, IJCV, CVPR Oral, ICCV, ECCV, NeurIPS Spotlight, ICML, ACM MM Oral, ICME Best Student Paper, DAI Best Paper, etc. He serves as a  Reviewer/PC in kinds of journals and conferences including TPAMI, IJCV,CVPR, ECCV, ICCV, NIPS etc.
\end{IEEEbiography}

\begin{IEEEbiography}[{\includegraphics[width=1in,height=1.25in,clip,keepaspectratio]{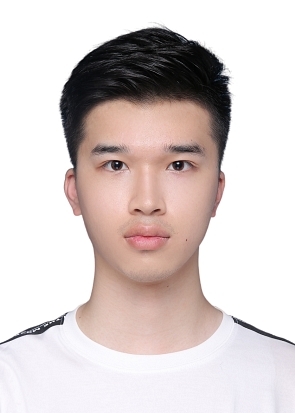}}]{Jiakang Yuan}
is currently a Ph.D. student in Electronic Engineering at the College of Future Information Technology, Fudan University (2022 - 2027).
Before that, he received his bachelor's degree in Electronic Engineering from Fudan University (2018 - 2022).
His research interests include multimodal reasoning, multi-agent system, and spatial intelligence. He has published papers in leading journals and conferences, including CVPR, ICCV, ECCV, NeurIPS, IEEE T-PAMI, etc, and serves as a reviewer in various journals and conferences, including IEEE T-IP, IEEE T-CSVT, CVPR, ECCV, ICCV, etc.
\end{IEEEbiography}
\begin{IEEEbiography}[{\includegraphics[width=1in,height=1.25in,clip,keepaspectratio]{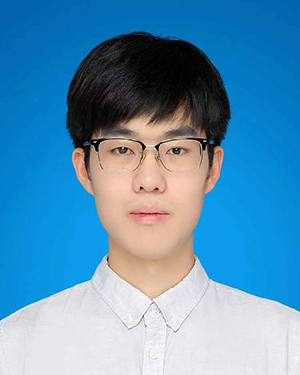}}]{Lin Zhang}
received the B.S. degree in Electronic Engineering from Fudan University, Shanghai, China, in 2022, where he is currently pursuing the Ph.D. degree with the College of Future Information Technology. His main research interests include computer vision and transfer learning.
\end{IEEEbiography}

\begin{IEEEbiography}[{\includegraphics[width=1in,height=1.25in,clip,keepaspectratio]{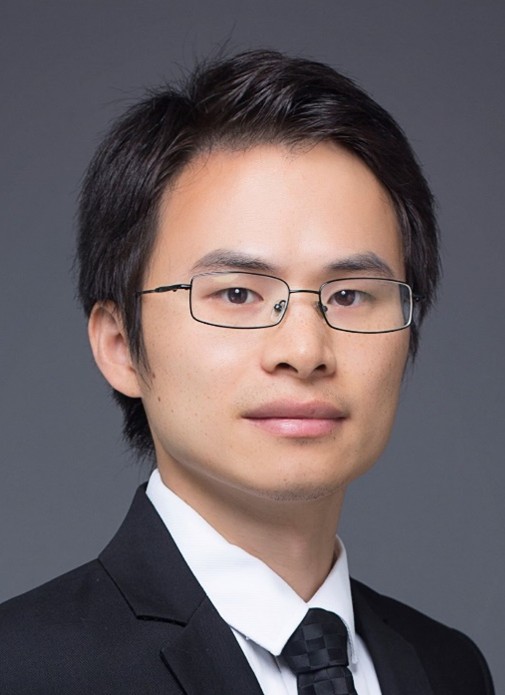}}]{Feng Xu}
(Senior Member, IEEE) received the B.E. degree (Hons.) in information engineering from Southeast University, Nanjing, China, in 2003, and the Ph.D. degree (Hons.) in electronic engineering from Fudan University, Shanghai, China, in 2008. From 2008 to 2010, he was a Post-Doctoral Fellow with the NOAA Center for Satellite Application and Research (STAR), Camp Springs, MD, USA. From 2010 to 2013, he was a Research Scientist with Intelligent Automation Inc., Rockville, MD, USA. Since 2013, he has been a Professor with the School of Information Science and Technology. He is also the Vice Dean of the School of Information Science and Technology and the director of the MoE Key Laboratory for Information Science of Electromagnetic Waves. He has published more than 100 articles in peer-reviewed journals and coauthored three books, among many conference papers and patents. His research interests include electromagnetic scattering theory, synthetic aperture radar (SAR) information retrieval, and intelligent radar systems.
Dr. Xu was the recipient of the 2023 Regional Leader Award and the 2014 Early Career Award of the IEEE Geoscience and Remote Sensing Society. Among other honors, he was awarded the second-class National Nature Science Award of China in 2011, the first-class Nature Science Award of Shanghai in 2022 and the first-class Technology Invention Award of CIE. He is currently the Director of Global Activities of IEEE GRSS. He also served as an Associate Editor for IEEE GEOSCIENCE AND REMOTE SENSING LETTERS (2014-2020) and for IEEE Transactions on Geoscience and Remote Sensing (2023-).

\end{IEEEbiography}

\begin{IEEEbiography}[{\includegraphics[width=1in,height=1.25in,clip,keepaspectratio]{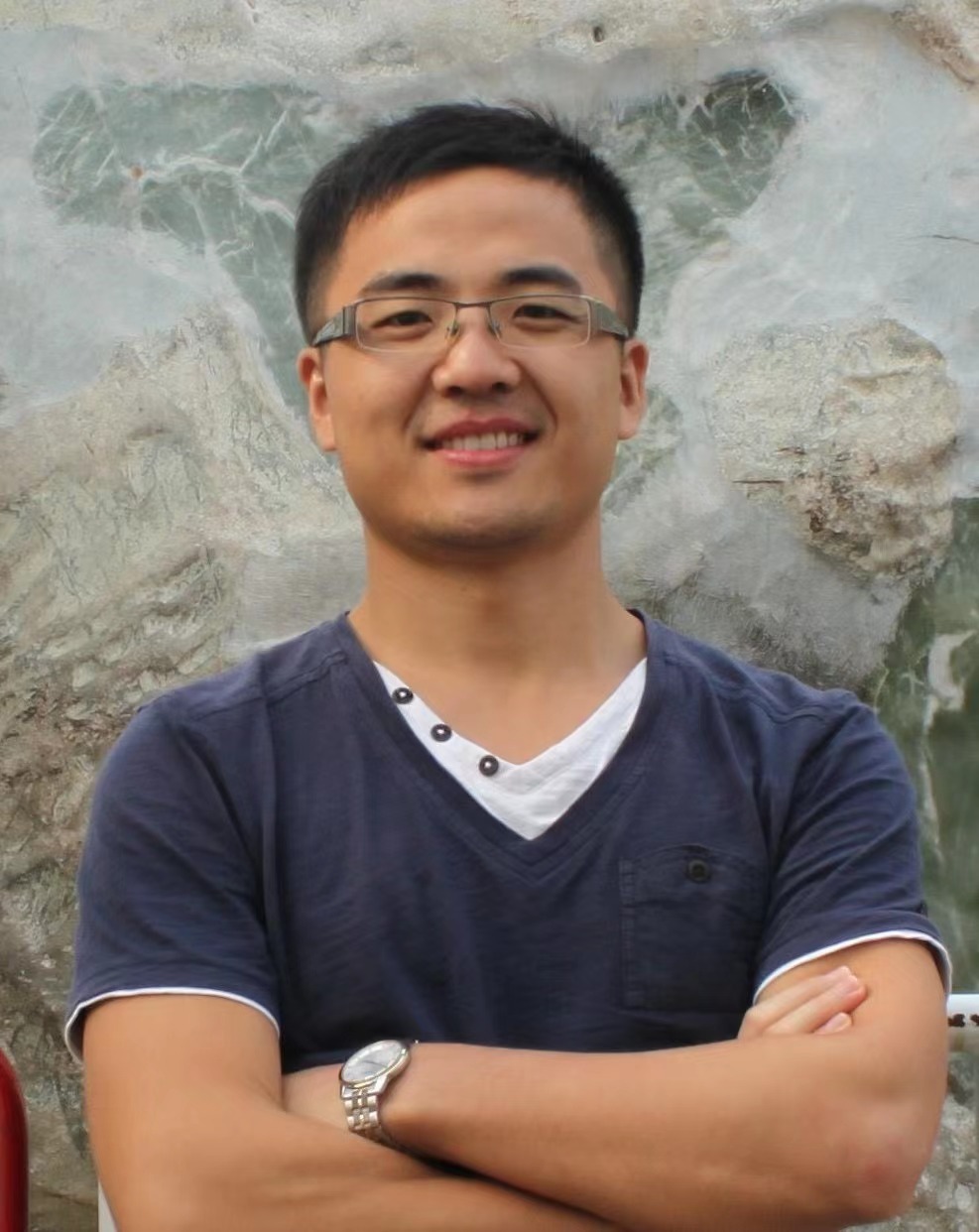}}]{Tao Chen}
(Senior Member, IEEE) received the PhD degree in information engineering from Nanyang Technological University, Singapore, in 2013. He was a research scientist with the Institute for Infocomm Research, A*STAR, Singapore, from 2013 to 2017, and a senior scientist with the Huawei Singapore Research Center from 2017 to 2018. He is currently a professor with the College of Future Information Technology, Fudan University, Shanghai, China. His main research interests include efficient computer vision, multimodal visual analysis, large VLM compression, and their applications in embodied robot intelligence, scene understanding and reconstruction. He has published over 200 top-tier papers in various reputable international journals and conferences such as IEEE T-PAMI, TIP, IJCV, CVPR, and NeurIPS. He has served as multiple ACs/SPC roles in conferences such as AAAI/ICLR/PRCV, etc. He also won the IJCAI 2025 Distinguished Paper Award, which was held in Montreal, Canada.
\end{IEEEbiography}

\end{document}



\bibliographystyle{IEEEtran}
\bibliography{references}

\end{document}